\documentclass{article}

 \usepackage[main, final]{neurips_2026}
\usepackage[utf8]{inputenc} % allow utf-8 input
\usepackage[T1]{fontenc}    % use 8-bit T1 fonts
\usepackage{hyperref}       % hyperlinks
\usepackage{url}            % simple URL typesetting
\usepackage{booktabs}       % professional-quality tables
\usepackage{amsfonts}       % blackboard math symbols
\usepackage{nicefrac}       % compact symbols for 1/2, etc.
\usepackage{microtype}      % microtypography
\usepackage{xcolor}         % colors
\usepackage{todonotes}

\usepackage{subfig}
\usepackage{threeparttable}
\usepackage{amsmath}
\usepackage{multirow}
\usepackage{booktabs} 
\usepackage{amssymb}
\usepackage{amsmath}
\usepackage{makecell}
\usepackage{pifont}
\usepackage{colortbl}
\usepackage{amsthm}
\usepackage{wrapfig}

\newtheorem{definition}{Definition}

\newtheorem{property}{Property}

\usepackage{tcolorbox}

\newtcolorbox{examplebox}{
  title=Example,
  colback=gray!5,
  colframe=gray!60
}

\title{Latent Debate: A Surrogate Framework for Interpreting LLM Thinking towards Binary Decisions}

\author{%
  Lihu Chen\textsuperscript{*} \qquad
  Xiang Yin\textsuperscript{* $\dagger$} \qquad
  Francesca Toni \\
  Imperial College London \\
  United Kingdom \\
  \texttt{\{lihu.chen,xy620,ft\}@imperial.ac.uk} \\
  \\
  {\small
  \textsuperscript{*}Equal contribution \quad
  \textsuperscript{$\dagger$}Corresponding author
  }
}

\newcommand{\ft}[1]{\noindent %\textcolor{orange!100}{(Francesca)} 
\textcolor{%orange!100
black}{#1}}

\newcommand{\newft}[1]{\noindent %\textcolor{orange!100}{(Francesca)} 
\textcolor{%orange!100
blue}{#1}}
\renewcommand{\newft}[1]{#1}

\definecolor{mypurple}{RGB}{128,0,128}
\newcommand{\lc}[1]{\textcolor{mypurple}{#1}}
\renewcommand{\lc}[1]{#1}

\begin{document}

\maketitle

\begin{abstract}
Understanding the internal `thinking' process of Large Language Models (LLMs) and %how this relates 
\newft{their connections} to hallucinations remains a key challenge. 
To this end, we introduce \emph{latent debate}, a structured surrogate framework for interpreting model outputs on True/False prediction tasks through the lens of %implicit 
internal latent arguments and interactions amongst them. Unlike %traditional explicit
human debates, latent debate captures the hidden supporting and attacking signals that arise within a %single 
model during a single inference. 
We first present a model- and task-agnostic conceptual framework, and then instantiate it symbolically to approximate the thinking process of LLMs %on True/False prediction tasks
towards binary decisions.
Empirical studies demonstrate that our latent debate is a faithful structured surrogate model that has highly consistent predictions with the original LLM, while providing a form of interpretability. We also  demonstrate that our latent debate provides a strong baseline for hallucination detection.
%Further  analysis reveals 
Specifically, we identify 
strong correlations between  debate patterns and hallucinations, such as a high degree of  
disagreement in the middle layers  of the latent debate surrogate is linked to a higher risk of hallucinations.
\lc{Our findings suggest that latent debate provides a new framework for understanding the internal processes of LLMs.} The code is available at \url{https://github.com/XiangYin2021/LatentDebate}.
\end{abstract}

\section{Introduction}
Large Language Models (LLMs) have %achieved
made remarkable progress on many reasoning tasks, yet they continue to suffer from hallucinations~\citep{xu2024hallucination, huang2025survey}.
For example, LLMs may generate answers that contradict user prompts or conflict with the source of training data~\citep{ji2023survey,kalai2025why,bang2025hallulens}, seriously undermining their reliability and trustworthiness.
This is further aggravated by the fact that, due to their opacity, it is difficult to understand why
LLMs make given predictions, or why their `thinking' process is flawed.

% \begin{figure}[t]  % r = wrap on the right; adjust width as needed
%   \centering
%   \subfloat[\textit{\textbf{Correct}: The city of Zhangzhou is in China.}]{%
%     \includegraphics[width=0.23\textwidth]{figures/true_heatmap.pdf}%
%   }\quad
%   \subfloat[\textbf{Hallucinated: }\textit{The letter J is the most commonly used letter in English-language writing.}]{%
%     \includegraphics[width=0.23\textwidth]{figures/false_heatmap.pdf}%
%   }%
%   \caption{
%   %\textcolor{}{ Concrete case study visualizing } 
%   Visualizations of 
%   %our 
%   latent debate for two claims (using %the last few 
%   selected tokens of Llama-8B)%in our framework
%   . Red/blue cells represent attacking/supporting arguments, respectively. %while blue cells represent supporting arguments. 
%   %More %debates tend to 
%   High 
%   %More 
%   controversy indicates hallucination. }%
%   \label{fig:teaser}%
% \end{figure}

Recent work in mechanistic interpretability has examined hallucinations through various internal signals, including activations~\citep{ferrando2025know}, attention patterns~\citep{chuang2024lookback}, and hidden states~\citep{azaria2023internal}. Another line of relevant research \citep{wang2022self} leverages external consistency, i.e., the agreement among multiple answers, to analyze hallucination behaviors. 
Their findings reveal that hallucinated outputs tend to have low self-consistency~\citep{wang2022self}.
This phenomenon suggests that strong agreement among multiple answers often yields more certain and accurate answers, whereas %greater debate 
disagreement indicates higher uncertainty 
and %a greater risk of hallucination
can serve as a good signal for detecting hallucinations.
%Subsequent 
Further approaches~\citep{irving2018ai, du2024improving, chen2024reconcile, liang2024encouraging} %further 
introduce multiple-agent debate %(MAD) 
to reduce hallucinated answers via a debate process of multiple %language models
LLMs, often outperforming single-model baselines.

Inspired by these prior studies of mechanistic interpretability and %external 
disagreement/debate, we aim to understand how hallucinations emerge within a model, but shifting %from explicit debate 
to a novel notion of \emph{latent debate}, i.e., %the 
%conflicts that arise
arising among different layers and %reasoning 
thinking steps within %the model
an individual model %and 
for a single inference ~\citep{chuang2023dola,liang2024internal,xie2024calibrating}, rather than externally %to it
as in prior work based on debate processes. %  
Unlike in conventional human debate,
where arguments are in natural language,
the arguments in our latent debate are to be understood in a metaphorical sense as they correspond to latent states in models. 
Intuitively, latent debate aggregates metaphorical arguments
to reflect the thinking and decision-making process beneath the surface.
In psychological theories, the human thinking process often involves internal debate-like behaviors such as inner speech~\citep{barker1966model} and the dialogical self~\citep{hermans2001dialogical}.
Here, we extend these psychological insights to describe an analogous process taking place within a model.

\begin{wrapfigure}{r}{0.5\textwidth}
  \centering
  \vspace{-10pt}
  
  \subfloat[\textit{\textbf{Correct}: The city of Zhangzhou is in China.}]{%
    \includegraphics[width=0.23\textwidth]{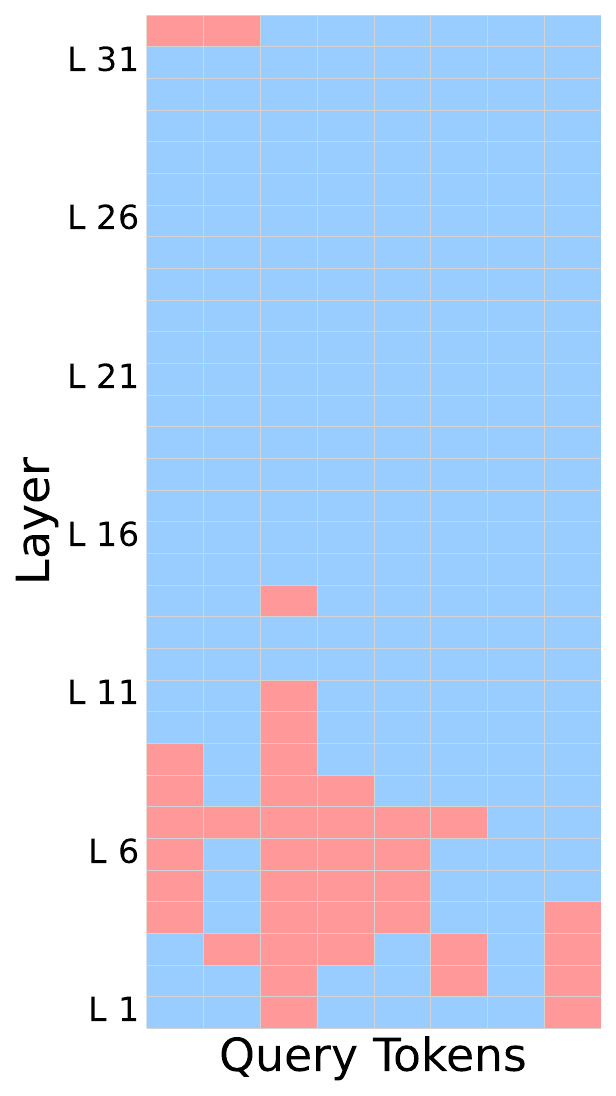}%
  }\quad
  \subfloat[\textbf{Hallucinated: }\textit{The letter J is the most commonly used letter in English-language writing.}]{%
    \includegraphics[width=0.23\textwidth]{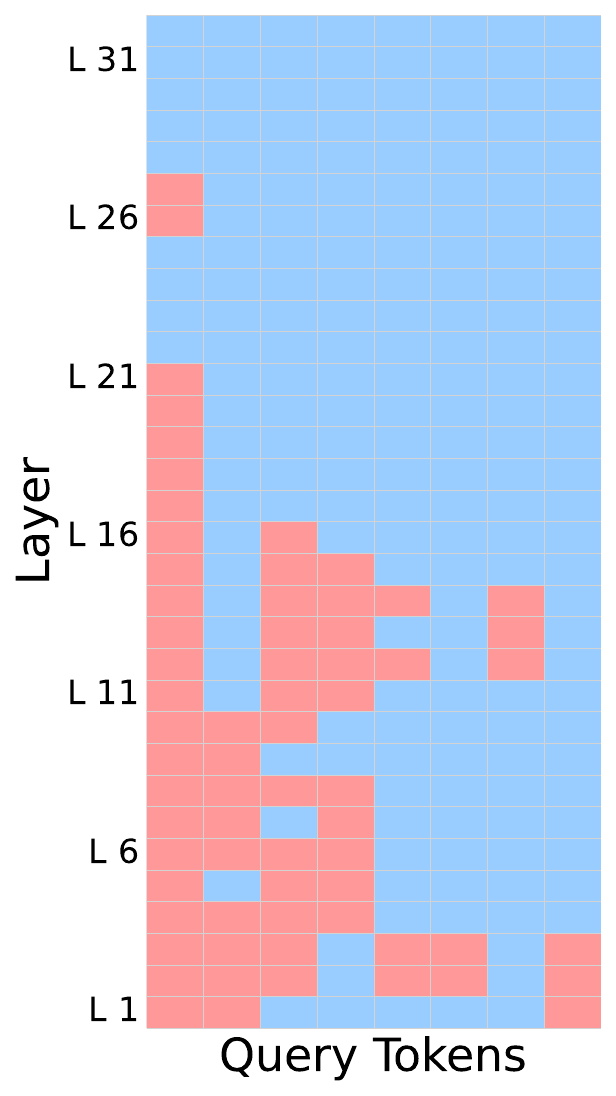}%
  }

  \caption{
  Visualizations of latent debate for two claims (both predicted as True) using selected tokens of Llama-8B. 
  Red/blue cells represent attacking/supporting arguments, respectively. 
  High controversy indicates hallucination.
  }
  
  \label{fig:teaser}
  \vspace{-10pt}
\end{wrapfigure}

We use the concept of \emph{latent debate} to build a transparent and simple surrogate model that approximates binary predictions of a complex black-box model, with the goal of understanding and interpreting the target model's behavior. 
We focus on two key research questions% in this work
: (1) \emph{Can we use our notion of latent debate to model the LLM thinking process? } (2) \emph{Can latent debate identify and interpret hallucinations?}
%In this work, we 
Thus, our approach is not intended to improve the target model's capabilities, such as improving factuality, but rather to provide insights into its internal decision-making process.

To answer the first question, we present a conceptual framework of latent debate that depicts  (dis)agreements within %a system
%an LLM
a model, while being model- and %\ft{(binary decision)} 
task-agnostic.   The framework consists of three abstract components: latent arguments derived from internal signals, an argument interpreter that translates these %implicit 
latent arguments into human-readable opinions, such as supporting or attacking a claim, and a thinking module that aggregates them to make the final decisions. We then instantiate this framework in decoder-based LLMs on True/False prediction tasks (see case studies in Figure~\ref{fig:teaser}), where hidden states serve as latent arguments, the unembedding matrix acts as the argument interpreter, and the thinking module is realized through a symbolic argumentation framework, in the spirit of \citep{vcyras2021argumentative}.
An empirical study demonstrates that this latent debate acts as a structured surrogate model, providing a faithful approximation of LLM thinking, which achieves up to 97\% consistency with LLaMA-13B decisions. 
These findings suggest that latent debate provides a useful approach to approximate internal signals for binary decision tasks.

To address the second question, we extract features from the latent debate% graph
, e.g., the number of internal %debates 
chains of disagreement and argument strengths, and train a simple %decision 
tree-based classifier to distinguish hallucinated from non-hallucinated outputs.
We find that our latent debate can achieve highly competitive performance in hallucination detection
for binary decision tasks.
We then use SHAP attribution scores \citep{SHAP} to identify which features most strongly drive hallucination predictions. Our analyses indicate that a high degree of controversy in the latent debate, particularly in the middle layers, is the strongest predictor of hallucination.

In summary, our contribution is threefold.
(1) We propose latent debate, a novel, model-agnostic framework that leverages internal arguments to approximate a model’s thinking process.
(2) We present a symbolic instantiation of latent debate that serves as a structured surrogate for LLM binary decision tasks.
(3) We develop a simple latent debate-based approach to detect hallucinations, which helps us identify distinct debate patterns.

\begin{figure*}
    \centering
    \scalebox{1.0}{
    \includegraphics[width=1.0\textwidth]{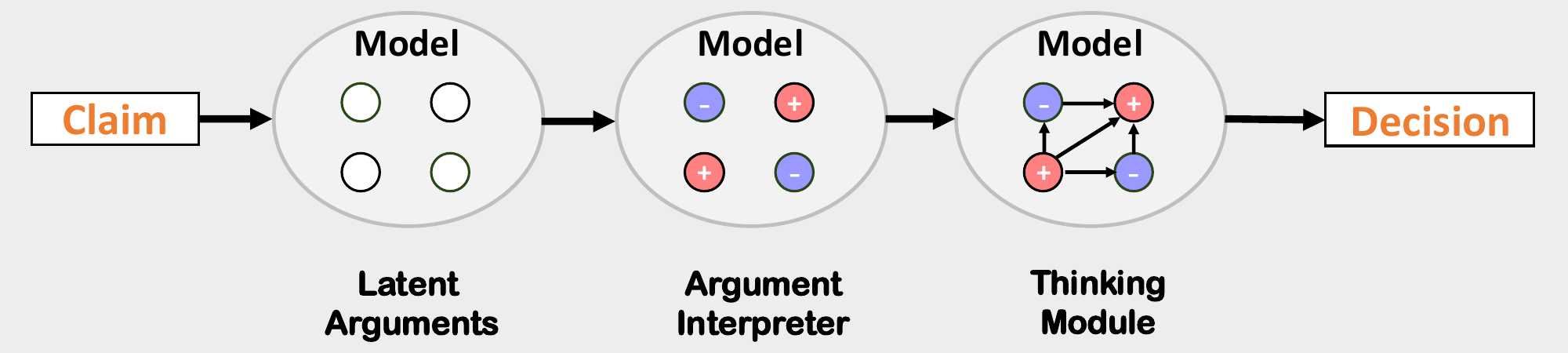}}
    \caption{The %overall 
    conceptual framework of latent debate. Given an input (e.g., a \emph{claim}), %our method generates 
    a set of \emph{latent arguments} \ft{are identified},
    i.e., model components (raw latent signals) that %implicitly 
%(metaphorically) 
convey the model's opinions toward the %claim
\ft{input}. These arguments are then processed by the \emph{argument interpreter},
identifying the arguments' supporting or attacking stance towards the %claim
\ft{input}.  The resulting
attacking and supporting arguments are fed into the \emph{thinking module}, which %applies a procedure to 
reaches the final \ft{True/False} \emph{decision}.}
    \label{fig:latent_debate_framework}
\vspace{-10pt}
\end{figure*}

\section{Related Work}
\paragraph{%Multiple Agent 
Debate \ft{in AI}.}
%Multiple-agent debate (MAD) 
\ft{Debate} has emerged as a powerful approach for improving factuality and reasoning. 
Pioneering work on AI safety via debate~\citep{irving2018ai} models debate as a self-play game with a (human) judge and provides core theoretical motivation.
Recent work adopts multiple %language model 
LLM agents to debate over individual responses jointly, with the final decision made either through consensus~\citep{du2024improving, chen2024reconcile} or by a judge~\citep{liang2024encouraging}. 
This debate strategy can outperform single model baselines on a wide variety of reasoning tasks. 
Subsequent research %continues this direction by proposing 
proposes refined debate approaches~\citep{li2024improving, liu2024groupdebate} using %MAD 
multi-agent debate as an evaluator~\citep{chanchateval}. In this work, we focus on debates operating implicitly within a model (agent) rather than externally visible debates among multiple agents.

\paragraph{Internal Consistency.}
Another line of work focuses on how to use the consistency of internal model states, such as logits and activations,  to improve model outputs~\citep{liang2024internal}. For example, DoLa~\citep{chuang2023dola} proposes a decoding strategy that contrasts logits between later layers and earlier layers for improving LLM truthfulness. \citet{xie2024calibrating} adopt internal consistency, i.e., how middle layers' predictions (dis)agree %(or disagree)
with the final layer, to guide LLM decoding.
In this work, we aim to use latent (dis)agreements to build a surrogate framework for interpreting the thinking process of a model rather than enhancing model outputs.
We also show 
that our surrogate model can help detect hallucinations.
%Thus our method and DoLa are orthogonal.

\paragraph{Computational Argumentation in Explainable AI.}

\emph{Argumentation Frameworks (AFs)}~\citep{Dung_95} are a fundamental formalism in \emph{computational argumentation}~\citep{CA-survey,vcyras2021argumentative}, a well-established research area in AI. According to~\citep{Dung_95}, an AF consists of a set of \emph{arguments} and a binary \emph{attack} relation among them. Arguments are %represented 
seen as abstract entities, while the attack relation captures conflicts between arguments.
AFs and their extensions, such as incorporating weights for arguments and a support relation between arguments, collectively referred to as Quantitative Bipolar Argumentation Frameworks (QBAFs)~\citep{rago2016discontinuity,baroni2019fine}, have been widely adopted in Explainable AI% (XAI)
\newft{~\citep{,vcyras2021argumentative}}.
AFs can serve as \emph{surrogate models} to approximate the inner structure and decision-making process of AI systems~\citep{vcyras2021argumentative,potyka2021interpreting, potyka2023explaining,ayoobi2023sparx}.
%Beyond serving as surrogates, AFs can also be explicitly integrated into AI or LLM systems to enhance explainability\citep{freedman2025argumentative,vcyras2021argumentative,vassiliades2021argumentation,engelmann2022argumentation,guo2023argumentative}. 
In this work, we are the first to introduce the model-agnostic concept of latent debate and adopt %an AF 
\newft{a QBAF for} %as 
the thinking module of %a model
\newft{an LLM}.

\section{Methodology}

\subsection{Problem Statement}~\label{sec:problem_statement}
%In this work, we %adopt the concept of latent debate to develop a
We 
aim to obtain a \emph{structured surrogate model} to understand the internal thinking of a target model,
as opposed to a \emph{conventional surrogate model} solely imitating the input-output behaviour  of the model.
More specifically, given a target model $M: \mathcal{X} \to \mathcal{Y}$, a conventional surrogate model $S: \mathcal{X} \to \mathcal{Y}$ %aims to imitate 
imitating 
the input-output behaviour of the %complex 
target model %, i.e., 
\ft{is such that} $S(x) \approx M(x)$. 
Such surrogates offer a simplified and interpretable approximation of the model's outputs~\citep{asher2015review,kudela2022recent}.
However, this type of surrogate does not truly reflect the internal structure (or internal thinking) of the model. 
To address this limitation, 
%we aim to develop a 
structured surrogate models
explicitly account for the model's internal organization~\citep{munk2022probabilistic,paez2024understanding}. 
%Supposing 
If the target model has a known internal 
structure: 
\begin{equation}
    M: \mathcal{X} \to \mathcal{H} \times \mathcal{Y}, \qquad  M(x) = (h(x), y(x))
\end{equation}
where $h(x) \in \mathcal{H}$ is the %internal 
structure, our goal is to construct% a surrogate
:
\begin{equation}
    S(x) = (\hat{h}(x), \hat{y}(x))
\end{equation}
that approximates the internal computational structure of the target model, i.e., \ft{such that} $\hat{h}(x) \approx h(x)$,
while, at the same time, %this surrogate should be 
being faithful to the target model by making highly consistent predictions with it, i.e., $\hat{y}(x) \approx y(x)$. Note that our main goal here is to build a simple surrogate model to imitate a complex model. Improving the ability of a model is not %the purpose of this work 
our purpose and is left to future work.

To obtain structured surrogate models $S(x)$
with the above characteristics,
we define the concept of \emph{latent debate} as follows.

\subsection{Conceptual Framework}

\paragraph{Latent Debate.} 
A latent debate is an internal %, implicit
form of argumentation that happens within a single model (or agent). Instead of having multiple explicit agents participating in a debate,  latent debate %refers to 
\ft{leverages upon} the hidden inconsistency inside the model that simultaneously carries supporting and attacking arguments toward %a claim
\ft{the input} \newft{to determine the corresponding output}. These arguments are not directly expressed in natural language but shape the model’s thinking process beneath the surface. 
The strength between supporters and attackers may be imbalanced. Overwhelming supporters can lead to a very certain positive decision, and vice versa.  
This uncertainty reflects how the model arrives at a final \ft{(binary)} decision. 
%The l
Latent debate consists of three key components: latent arguments, argument interpreter, and thinking module, as shown in Figure~\ref{fig:latent_debate_framework}.

\paragraph{Latent Arguments.} 
A latent argument %refers to 
is 
an internal signal within a model, implicitly conveying supporting or attacking opinions toward %a claim
\ft{the input}. 
Such signals can %derive 
arise from different sources, %like
e.g., activations or attention patterns. %Because 
Since they live in the model’s latent space, these arguments are not directly visible or human-readable, but they still express how intermediate steps think about the %claim
\ft{input}.

\paragraph{Argument Interpreter.}
The argument interpreter %is the tool that 
makes these latent arguments interpretable. It translates latent arguments into a form we can understand, such as a binary label.
At the same time, it tells us how strongly each argument supports or attacks the %claim
\ft{input}, turning vague internal signals into measurable opinions.

\paragraph{Thinking Module.}
The thinking module combines all the decoded arguments to reach a final decision. It looks at how the arguments interact — some supporting, some attacking — and weighs them against each other. By aggregating these arguments and how they %debate
interact, the module produces a final outcome that reflects the overall internal debate of the model.

It is important to note that this framework is not tied to any specific model architecture. The notions of latent arguments, argument interpreter, and thinking module are abstract components that can be realized in many different ways.
For example, latent arguments may be instantiated through hidden states, attention patterns, or other internal signals; argument interpreters can be designed using projection, probing, or alternative interpretability tools; and thinking modules may adopt symbolic argumentation frameworks, probabilistic aggregation, machine-learning methods, 
including artificial neural networks. This flexibility ensures that the latent debate framework can be adapted to a wide range of models and tasks beyond the particular instantiation we study next in this paper.

\subsection{Symbolic Instantiation for Latent Debate in LLM True/False Prediction %Tasks
}~\label{sec:instantiation}
% \subsubsection{Instantiation}
We now describe how the abstract concepts of latent debate can be instantiated in the context of %transformer-based 
LLM True/False prediction tasks~\citep{vaswani2017attention}. 
We adopt a symbolic argumentation framework 
%to perform the decision making process
\ft{as the thinking module},  which is transparent and efficient.

Formally,  given %a query 
\ft{an input} $\mathbf{x} = (x_1, \ldots, x_N)$ with a binary label $c \in \{\text{True}, \text{False}\}$, a decoder-based LLM generates %an %answer 
\ft{a decision} 
$\mathbf{y}= (y_1, \ldots, y_T)$. 
Both the %query 
\ft{input} and %answer 
\ft{decision} tokens are drawn from the same vocabulary, i.e.,
$x_n, y_t \in \mathcal{V}$.
Each token $y_{t}$ in the %answer 
\ft{decision} is generated conditionally based on the preceding tokens and the input% query
, following the distribution: $y_{t} \sim P(y_{t} \mid \mathbf{y}_{\leq t-1}, \mathbf{x})$. 
In a multiple agent debate
%MAD 
setting~\citep{du2024improving,liang2024encouraging}, the process involves generating multiple %answers
\ft{decisions} $\mathcal{Y} = \{ \mathbf{y}^{(1)}, \ldots, \mathbf{y}^{(K)} \}$, which may support ($\mathbf{y}^{+}$) or attack ($\mathbf{y}^{-}$) the %claim
\ft{input}. A final decision is then derived by aggregating these arguments, often through some form of consensus or voting strategies.  In contrast, we define that a latent debate takes place inside the model processing  %claim 
$\mathbf{x}$ before generating %answer 
$\mathbf{y}$.

\subsubsection{Latent Arguments in LLMs} 
An LLM consists of $L$ layers. Let $f_{\theta}$ denote the transformation function for computing hidden states, parameterized by $\theta$. 
The hidden state for the token $x_n$ of the %claim 
\ft{input} at layer 
$l$ is computed as: 
$
    \mathbf{h}_{n}^{(l)} = f_{\theta}(\mathbf{h}_{1}^{(l-1)}, \ldots, \mathbf{h}_{n}^{(l-1)})
$,
where $\mathbf{h} \in \mathbb{R}^{d}$,  with $d$ the dimensionality of the hidden states, corresponds to the normalized sum of residual and sub-layer outputs.
We treat each hidden state $\mathbf{h}_{n}^{(l)}$ as a \emph{latent argument}, a representation that implicitly encodes supportive or attacking views with respect to the %claim
\ft{input}, though not directly observable in natural language. 
Given a claim comprising $N$ tokens and an LLM with $L$ layers, we thus obtain  $N \times (L-1)$ latent arguments over the %network
LLM’s internal computation.
We exclude the final (output) layer because it directly produces the probability distribution over next tokens.
Instead, our goal is to depict the intermediate thinking dynamics encoded in the hidden layers prior to that final mapping.

\subsubsection{Argument Interpreter in LLMs}
To make latent arguments interpretable, %we introduce an 
the instantiated argument interpreter %. Specifically, it 
projects a hidden state into the vocabulary space using the unembedding matrix $\mathbf{W}^{\text{unemb}} \in \mathbb{R}^{|\mathcal{V}| \times d}$.
This produces a probability distribution over vocabulary tokens, which has been widely used in mechanistic studies~\citep{nostalgebraist2020interpretinggpt,belrose2023eliciting}. By examining the probabilities assigned to specific tokens \texttt{True} and \texttt{False}, we can quantify the opinion of each latent argument, i.e., how much it supports or attacks the %claim
\ft{input}.

\begin{equation}
    \label{equ_softmax}
    \text{interpret}(\mathbf{h}_{n}^{(l)}) = \text{Softmax}(\mathbf{W}_{[\text{True, False}]}^{\text{unemb}}(\mathbf{h}_{n}^{(l)}))
\end{equation}

The output of %the function 
%\textit
{interpret}($\cdot$) enables interpretation of the latent arguments through the lens of token-level semantics. 

\subsubsection{Thinking Module in LLMs}
Finally, a thinking module is applied to the set of interpretable arguments in order to produce a final judgement 
$c \in \{\texttt{True}, \texttt{False}\}$. This process is formalized as:

\begin{align}
\label{equ_probability}
\text{cls}(\mathbf{x}) &= \text{think}(\mathcal{H}), \qquad
\mathcal{H} = \left\{ \text{interpret}\!\left(\mathbf{h}_n^{(l)}\right)
\;\middle|\; 1 \leq n \leq N,\ 1 \leq l \leq L-1 \right\}
\end{align}

where $\text{cls}(\mathbf{x})$ is capable of outputting a label associated with the final decision $c$. To perform the thinking step, we %adopt 
\ft{adapt} a symbolic approach \ft{from computational argumentation}, Quantitative Bipolar Argumentation Framework (QBAF)~\citep{baroni2019fine}, as the $ \textrm{think}(\cdot)$ function, which accounts for both supporting and attacking relationships among arguments to yield a coherent, weighted judgment.

\begin{figure*}[!t]%
\centering
\subfloat[\centering 
%Graphical view of (elements of) an 
An example QBAF]{{\includegraphics[width=0.53\textwidth]{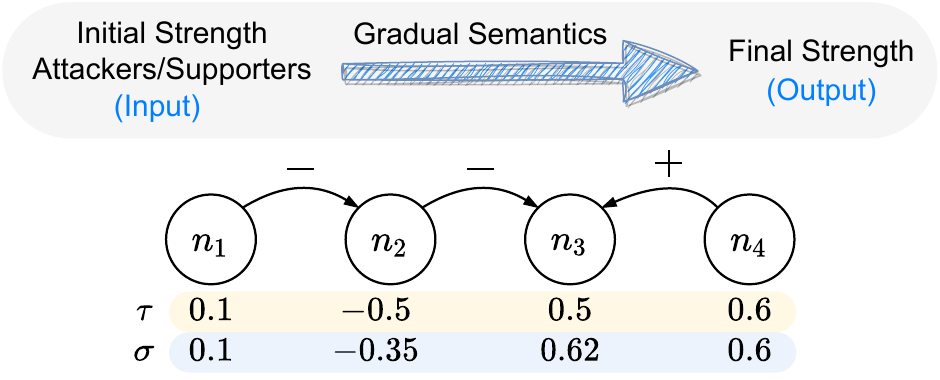}} \label{fig_semantics}}
\subfloat[\centering  %Visualizations of 
QBAF skeleton for LLMs.]{{\includegraphics[width=0.35\textwidth]{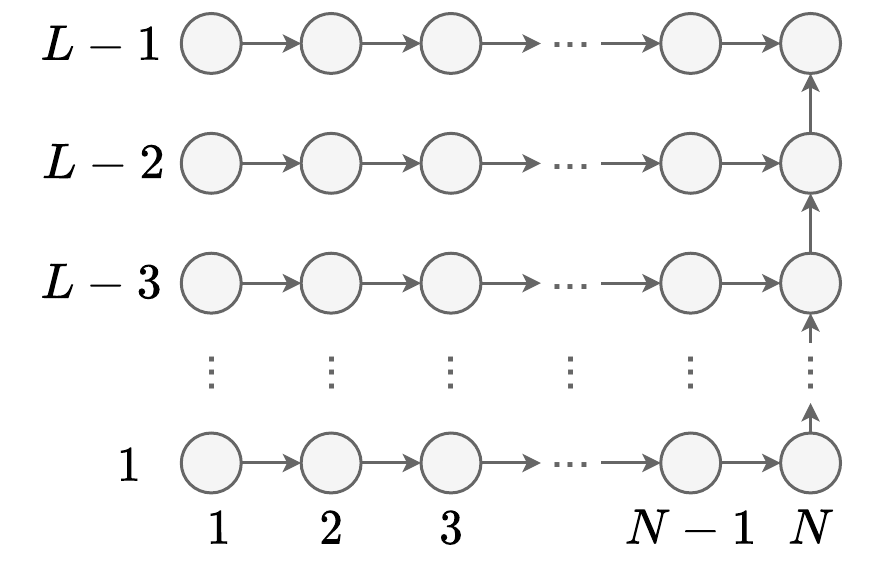} }\label{fig:qbaf_framework}}%
\caption{
%An overview of our QBAF and the instantiated graph for LLM architectures.
QBAFs and LLMs. (a) An example QBAF showing how initial  strengths ($\tau$) of arguments $n_1, \ldots, n_4$ are transformed through gradual semantics based on attacking (-) and supporting (+) relations to produce final strengths ($\sigma$).
(b) %The corresponding 
Skeleton of a QBAF %when applied to LLM architectures
drawn from an LLM architecture, where each node represents a specific token at a specific layer. To obtain a QBAF, the directed edges %depict how to create the argumentation graph
need to become attacks or support and the nodes need to be equipped with an initial strength.}
\label{fig:correct_incorrect_fraction}%
    \vspace{-10pt}
\end{figure*}

\begin{definition}%[QBAF]
\label{def_QBAF}
A \emph{QBAF} is a quadruple $\mathcal{Q}=\left\langle\mathcal{A}, \mathcal{R}^{-}, \mathcal{R}^{+}, \tau \right\rangle$ where $\mathcal{A}$ is a finite set of \emph{arguments}; $\mathcal{R}^{-} \subseteq \mathcal{A} \times \mathcal{A}$ is a binary \emph{attack} relation;
$\mathcal{R}^{+} \subseteq \mathcal{A} \times \mathcal{A}$ is a binary \emph{support} relation;
$\tau$ is an \emph{initial strength function} ($\tau: \mathcal{A} \rightarrow  [-1,1]$).
\end{definition}

%Given a set of arguments $\mathcal{A}$, a 
A \ft{gradual semantics~\citep{baroni2015automatic,rago2016discontinuity,Potyka18,amgoud2018evaluation} applied to a} QBAF is capable of considering the overall debate situation and outputting \emph{final strengths} \ft{for all arguments} (as shown in Figure~\ref{fig_semantics}), which can be used to obtain the binary (\textit{True/False}) predictions \ft{(e.g. by thresholding)}. 
In the figure, each node $n_i$ corresponds to an argument, and the $\tau(\cdot)$ function assigns the \emph{initial strength},  indicating its polarity and strength magnitude before %propagation
\ft{applying the semantics}. The directed edges between nodes represent the relationships% among arguments
: edges labeled with $-$ are \emph{attacks}, indicating that one argument undermines another (two arguments with different polarities), while edges labeled with $+$ are \emph{supports}, meaning that one argument reinforces another. The $\sigma(\cdot)$ beneath the nodes is \ft{the gradual semantics},% a function resulting final strengths after applying a chosen gradual semantics, e.g., \citep{baroni2015automatic,rago2016discontinuity,Potyka18,amgoud2018evaluation},
which reflects how the collective influence of attackers and supporters modifies the outcome. See Example~\ref{example_semantics} in the appendix for a detailed computation process.

Formally, gradual semantics provide the rules that control how initial strengths are updated and result in the computation of \emph{final strengths}.
%For this goal, 
\ft{To match the thinking of LLMs, we} define a new semantics. We start from arguments with no attackers or supporters, whose final strength is the same as their initial strength. 
For the remaining arguments, the final strength is updated along the edges by considering the influence of both attackers and supporters. \ft{As conventional~\citep{potyka2021interpreting},} this process involves two components: \emph{aggregation} and \emph{influence}.
For an argument $\alpha \!\!\in\!\! \mathcal{A}$, the aggregation step computes its \emph{energy} $\!E_\alpha\!$ by summing the strengths of its attackers and supporters ($\beta$):
\begin{equation}
    E_\alpha=\sum_{\left \{ \beta \in \mathcal{A} \mid (\beta,\alpha) \in \mathcal{R^{-} \cup R^{+}} \right \} }\sigma(\beta)
\end{equation}
The influence step then updates the initial strength $\tau(\alpha)$ by combining it with the computed energy:
\begin{equation}
\sigma(\alpha) = \tanh\left(E_\alpha \right) + \tau(\alpha) \cdot \left(1 - \tanh\left(\left|E_\alpha\right|\right)\right)
\end{equation}
This equation updates the final strength of an argument by combining the aggregated influence from its attackers and supporters ($\tanh(E_\alpha)$) with its own initial strength $\tau(\alpha)$. \lc{Here, $\tau(\alpha)=w \times \text{interpret}(\mathbf{h}_{n}^{(l)})$}, 
where $w$ is the token-wise weight that measures the semantic contribution of this current token to the entire sentence (see details in Section~\ref{app:token_wise_weight}). 
In other words, the weight scales how strongly the initial strength %\todo{what? lacks a verb} 
\ft{propagates} since each token contributes in a different way to the semantic meaning. 

The strength function $\sigma: \mathcal{A} \rightarrow [-1,1]$ assigns each argument a value where the sign indicates polarity \ft{towards the output decision} (supportive or attacking) and the absolute value indicates magnitude.
The values in [-1,0) correspond to negative labels, while values in [0,1] correspond to positive labels.

\paragraph{Creating QBAFs for LLMs}
Figure~\ref{fig:qbaf_framework} illustrates how we construct a QBAF for LLM architectures. Each row corresponds to a transformer layer, and each circle represents an argument associated with a thinking step (\textit{a token}) at that layer.
In our instantiation,we treat the last few tokens of the prompt as thinking steps, which are the tokens  
generated after the model has already  seen the entire %question
\ft{input}.
In this work, we treat the final token of the input% question
, along with the subsequent auxiliary tokens (\textit{``The statement is True or False:''}), as the model's thinking steps.
In this way, even though thinking tokens cannot attend to subsequent tokens, they can attend to the full question and the beginning prompt, which is sufficient to provide the task
specification and the basic context. 
Hence, the tokens after the %question
\ft{input} serve as meaningful intermediate thinking units in the latent debate process. 

Arguments are first connected within a layer from left to right, following the natural order of tokens in the input sequence. The right-most node in each layer summarizes the thinking results for that layer. We then connect these right-most nodes across layers, from lower to higher, since upper layers are closer to the final decision. The node in the top-right corner thus considers the overall information to make decisions, and we use its output as 
%this binary classifier.
\ft{the binary classification}.
The initial strength of each argument is determined by its probability defined in Equation~\ref{equ_softmax}, which is then normalized to a value in $[-1, 1]$. The sign of the initial strength reflects its polarity (positive or negative), while the magnitude encodes the confidence. Relations between arguments are determined by comparing polarities: if two connected arguments share the same polarity, the edge is a support. Otherwise, it is an attack. Because polarity depends on evolving strengths during computation, edge types may be updated dynamically.

This topology is intentionally simple to enhance %explainability
\ft{interpretability}. In particular, we avoid connecting arguments of the same token across layers as this way does not bring clear benefits (see results of quadratic connections in Table~\ref{tab:correlation}).
In the current implementation, our framework captures structural patterns within the internal structure that are relevant for decision-making rather than merely input-output imitation, i.e.,
\ft{latent debate is} \emph{a structured surrogate}.

\subsubsection{Benefits}

\textbf{(1) Transparent and Interpretable.} 
Our framework makes the internal thinking process of LLMs human-readable via a symbolic argumentation framework. Each latent signal is translated to a clear supporting or attacking argument, and the QBAF decision path can be visualized and explained rather than remaining a black box.
\textbf{(2) Training-Free.} 
Our framework %works\todo{delete "works"?} 
imitates the LLM thinking process without any tuning and training% samples
. Every component is lightweight, which makes the method easy to use.
\textbf{(3) Property-Satisfying.} 
Because the reasoning process is formalized with a symbolic argumentation framework, the method inherits desirable theoretical properties such as monotonicity in \citep{baroni2018many}. In practice, this guarantees that the framework behaves in an intuitively consistent way when adding new arguments or changing the initial strength of arguments (see the proof and details in the appendix~\ref{app:property}).

\begin{table*}[tp]
    \centering
    \tiny
    \setlength{\tabcolsep}{2.2mm}
    \begin{threeparttable}
        \begin{tabular}{lccccccc}
            \toprule
            & \textbf{\underline{\texttt{common\_claim}}}
            & \textbf{\underline{\texttt{counterfact}}}
            & \textbf{\underline{\texttt{company}}}
            & \textbf{\underline{\texttt{TriviaQA}}}
            & \textbf{\underline{\texttt{MuSiQue}}}
            & \textbf{\underline{\texttt{TruthfulQA}}}
            & \textbf{\underline{\texttt{Avg}}} \\
            & 500 & 500 & 500 & 500 & 500 & 500 & -- \\
            \midrule
            \multicolumn{8}{c}{\textit{Llama-8B (\%)}} \\
            \midrule
            Random
            & 76.8 & 64.2 & 67.4 & 66.4 & 68.8 & 79.2 & 70.5 \\
            Average
            & 92.2 & 67.4 & 80.4 & 73.0 & \textbf{77.0} & 90.2 & 80.0 \\
            Majority Voting
            & \textbf{92.4} & 67.0 & 80.4 & 73.0 & \textbf{77.0} & 90.2 & 80.0 \\
            \rowcolor{gray!10}
            Latent Debate
            & \textbf{92.4} & \textbf{78.2} & \textbf{89.2}
            & \textbf{74.0} & \textbf{77.0} & \textbf{90.6}
            & \textbf{83.6} \\
            Latent Debate -- w/o token weight
            & 92.2 & 68.2 & 80.2 & 73.0 & \textbf{77.0} & 90.2 & 80.1 \\
            Latent Debate -- with quadratic connection
            & 91.8 & 64.0 & 80.0 & 73.2 & \textbf{77.0} & 90.2 & 79.4 \\
            \midrule
            \multicolumn{8}{c}{\textit{Mistral-7B (\%)}} \\
            \midrule
            Random
            & 64.0 & 65.4 & 74.0 & 75.0 & 75.8 & 70.4 & 70.8 \\
            Average
            & 86.8 & 89.0 & \textbf{98.8} & 95.6 & 90.2 & 84.2 & 90.8 \\
            Majority Voting
            & 89.8 & 90.6 & 98.6 & \textbf{96.0}
            & \textbf{91.6} & 87.0 & 92.3 \\
            \rowcolor{gray!10}
            Latent Debate
            & \textbf{90.0} & \textbf{91.0} & 97.8
            & 95.4 & 91.2 & \textbf{89.2} & \textbf{92.4} \\
            Latent Debate -- w/o token weight
            & 87.0 & 89.0 & \textbf{98.8} & 95.6 & 90.2 & 84.0 & 90.8 \\
            Latent Debate -- with quadratic connection
            & 81.2 & 76.8 & 91.6 & 95.0 & 90.2 & 77.6 & 85.4 \\
            \midrule
            \multicolumn{8}{c}{\textit{Llama-13B (\%)}} \\
            \midrule
            Random
            & 65.2 & 65.0 & 78.8 & 66.8 & 62.4 & 68.8 & 67.8 \\
            Average
            & 90.0 & 90.0 & 98.8 & 86.2 & 85.4 & 89.6 & 90.0 \\
            Majority Voting
            & 85.4 & 88.0 & 98.2 & 84.2 & 84.4 & 87.0 & 87.9 \\
            \rowcolor{gray!10}
            Latent Debate
            & \textbf{98.4} & \textbf{95.2} & \textbf{99.6}
            & 96.2 & \textbf{93.6} & \textbf{96.8} & \textbf{96.6} \\
            Latent Debate -- w/o token weight
            & 95.6 & 91.2 & 98.6 & \textbf{97.6} & 93.2 & 93.6 & 95.0 \\
            Latent Debate -- with quadratic connection
            & 80.2 & 66.0 & 59.8 & 95.2 & 93.0 & 83.4 & 79.6 \\
            \bottomrule
        \end{tabular}
        \caption{Consistency scores on binary and open-ended
        question-answering datasets with three LLMs.
        Each dataset contains 500 examples.
        The first three datasets use the last six prompt tokens,
        while TriviaQA, MuSiQue, and TruthfulQA use the last five.
        Avg is the arithmetic mean across the six datasets.
        The best scores within each model are shown in bold.}
        \label{tab:correlation}
    \end{threeparttable}
    \vspace{-10pt}
\end{table*}

\section{Latent Debate as a Surrogate for Imitating LLM True/False Prediction}\label{sec:approximate_decision_making}

A core motivation of our latent debate is to approximate the thinking process of LLMs, %which allows 
allowing us to interpret their internal mechanisms. 
We %hope 
\ft{wish that} this transparent framework is faithful to the original model's predictions (\textit{Prediction Fidelity}), i.e., the surrogate model should match outputs of the target black-box model~\citep{papenmeier2019model,laugel2018defining}.
To validate this, we conduct experiments on three balanced True/False prediction datasets: \textit{common claims, counterfact, and company}. 
We also include three open-ended question-answering datasets mapped onto binary claims: \textit{TriviaQA, MuSiQue, and TruthfulQA} (see details in section~\ref{app:dataset}).
We use three open-
weight LLMs in the experiments: \textit{meta-llama/Llama-3.1-8B, mistralai/Mistral-7B-Instruct-v0.3, and meta-llama/Llama-2-13B.}

To benchmark the faithfulness of our latent debate approach, we compare it against several intuitive baselines of \emph{structured surrogates}~\footnote{In this experiment, non-structured surrogate baselines are not compared, e.g., single-argument baselines (see discussions in Section~\ref{sec:single_arg}). }: (1) \textit{Random}, (2) \textit{Average}, (3) \textit{Majority Voting}, (4) \textit{Latent Debate -- w/o token weight}, and (5) \textit{Latent Debate -- with quadratic connection}. The details of baselines are described in Section~\ref{app:baseline_surrogate}. 
Since we aim to propose a transparent surrogate model, approaches that employ internal signals to improve output quality, such as DoLA~\cite{chuang2023dola}, are not included in this experiment.
We report the \emph{consistency score}, i.e., the proportion of instances on which the decision derived via the latent debate exactly matches the original LLM's prediction.

Table~\ref{tab:correlation} reports the consistency scores of different methods over 500 examples per dataset and three models. The latent debate approach achieves higher average consistency than baseline methods (\textit{Random, Average, Majority Voting}) and two variants of latent debate for all three models, while matching or slightly trailing a baseline on a few individual datasets. For instance, with Llama-13B the latent debate method reaches 96.6\% average consistency, while the best non-debate baseline (\textit{Average}) reaches 90.0\%. This demonstrates that latent debate is a strong structured surrogate for the True/False decisions made by the model. 
More importantly, our transparent approach %remains 
\ft{gives} interpretability by displaying internal supporting vs.\ attacking arguments that can be visualized and understood.

\section{Can Latent Debate Identify Hallucinations?}
Given that our latent debate surrogate aligns closely with the LLM's internal decision behavior, 
we now turn to %an interesting and 
a more diagnostic question: \emph{Can we use latent debate to detect hallucinated answers?}
Here, the term `hallucination' refers to their (lack of) factuality, in the spirit of~\citep{huang2025survey,zhang2025siren}, i.e., hallucinations emerge when the LLM's answers are inconsistent with established world knowledge.
In other words, we use our transparent and faithful surrogate model to detect and analyze how the model thinks and why it hallucinates. 

\subsection{Latent Debate is a Strong Baseline for Hallucination Detection}\label{sec:hallu_detector}

In order to detect hallucinations, we train a simple tree-based classifier, XGBoost~\cite{chen2016xgboost}, to distinguish hallucinated from non-hallucinated outputs using features extracted from latent debate, which we refer to as \emph{Latent Debate Detector}. 
The details of this detector are described in Section~\ref{sec:mlp} in the appendix.
This pipeline allows us to both detect hallucination and interpret why they happen in terms of internal debate patterns.  
% It is worth noting that we can also use other types of classifiers, but logistic regression is unfeasible in this experiment since it assumes a linear relationship between features, which makes it unfeasible to capture nonlinear and U-shaped interaction effects among our defined features~\citep{ranganathan2017common}.

Concretely, we extract the following five features related to debate patterns from each QBAF %in a language model
\ft{drawn from an LLM}: 
(1) \textit{number of attacks (NumAtk)}: the total number of attack edges in the QBAF, capturing how many conflicting arguments are present.
(2) \textit{average of initial strength (AvgInit)}: the arithmetic mean of the raw strengths of all latent arguments in the QBAF.
(3) \textit{average of final strength (AvgFin)}: the arithmetic mean of strength values after propagation under the chosen gradual semantics. (4) \textit{variance of initial strength (VarInit)}: the statistical variance of the raw strengths of all latent arguments in the QBAF.
(5) \textit{variance of final strength (VarFin)}: the statistical variance of strength values after propagation under the chosen gradual semantics. 
Since not all layers contribute equally to a model's reasoning process, we divide latent arguments evenly across layers into three regions for better understanding internal mechanisms: lower, middle, and upper. 
We extract the above five features for each region.

\begin{table*}[t]
    \centering
    \tiny
    \setlength{\tabcolsep}{2.0mm}{
        \begin{threeparttable}
            \begin{tabular}{lccccccc}
                \toprule
                \textbf{Method}
                & \textbf{Common Claim}
                & \textbf{CounterFact}
                & \textbf{Company}
                & \textbf{TriviaQA}
                & \textbf{MuSiQue}
                & \textbf{TruthfulQA}
                & \textbf{Avg} \\
                \midrule

                Hallucination Ratio
                & 40.8\% & 15.8\% & 30.2\%
                & 25.0\% & 35.2\% & 45.8\% & 32.1\% \\
                \midrule

                AvgInit
                & 0.75 & 0.66 & 0.86 & 0.91 & 0.71 & 0.58 & 0.74 \\

                AvgFin
                & 0.58 & 0.54 & 0.68 & 0.82 & 0.62 & 0.62 & 0.64 \\

                VarInit
                & 0.82 & 0.70 & 0.89 & 0.91 & 0.78 & 0.54 & 0.77 \\

                VarFin
                & 0.78 & 0.63 & 0.78 & 0.90 & 0.65 & 0.50 & 0.71 \\

                NumAtk
                & 0.55 & 0.53 & 0.68 & 0.65 & 0.51 & 0.55 & 0.58 \\

                SelfCheckGPT
                & 0.60 & 0.53 & 0.72 & 0.65 & 0.58 & 0.45 & 0.59 \\

                Semantic Entropy
                & $0.72 \pm 0.00$
                & $0.69 \pm 0.01$
                & $0.71 \pm 0.01$
                & $0.77 \pm 0.01$
                & $0.52 \pm 0.00$
                & $0.51 \pm 0.01$
                & 0.65 \\

                SAPLMA
                & $0.82 \pm 0.01$
                & $0.77 \pm 0.00$
                & $0.86 \pm 0.00$
                & $0.90 \pm 0.01$
                & $\mathbf{0.85 \pm 0.01}$
                & $\mathbf{0.71 \pm 0.03}$
                & \textbf{0.82} \\

                \rowcolor{gray!10}
                Latent Debate Detector
                & $\mathbf{0.84 \pm 0.00}$
                & $\mathbf{0.83 \pm 0.00}$
                & $\mathbf{0.93 \pm 0.00}$
                & $\mathbf{0.91 \pm 0.00}$
                & $0.78 \pm 0.01$
                & $0.61 \pm 0.00$
                & \textbf{0.82} \\

                \bottomrule
            \end{tabular}

            \caption{
                \lc{AUROC scores for the identified features in isolation
                (AvgInit, \ldots, NumAtk), hallucination detection
                baselines, and Latent Debate Detector.
                Results are reported as the average score
                over three random seeds where available.
                The best results are highlighted in bold.}
            }
            \label{tab:mlp_auc}
        \end{threeparttable}
    }
    \vspace{-10pt}
\end{table*}

We compare our latent debate detector to the value of these five features in isolation and another two commonly-used methods for hallucination detection: SelfCheckGPT~\citep{manakul2023selfcheckgpt} and SAPLMA~\citep{azaria2023internal}.  
The baseline details are described in Section~\ref{sec:baseline_hallucination} in the appendix.
% \textcolor{red}{We also considered DoLa as an additional baseline, given its use of internal agreement, but opted against it because it uses this agreement to obtain a stand-alone model rather than as a surrogate model. Lihu: move this to section 5.1  } 
Table~\ref{tab:mlp_auc} shows the comparison across different hallucination detectors.
We can see that our latent debate detector can achieve a decent AUROC compared to baselines on average, which suggests that our approach can serve as strong baseline in distinguishing hallucinations from non-hallucinations.
More importantly, our simple method offers interpretable features to analyze why the model hallucinates, %which we will discuss in the next section. 
\ft{discussed next}.

\subsection{What Debate Patterns %Cause
are Correlated with Hallucinations?}\label{sec:more_debate_hallucination}%\todo{cause->are associated with?}
To study which debate pattern is correlated with hallucinations, we apply SHAP attribution~\citep{SHAP} to determine which features most strongly contribute to hallucination. Using 
%Decision Trees
\ft{tree-based classifiers} with SHAP attribution is a broadly adopted approach for feature analysis and interpretability~\citep{lundberg_unified_2017,ponce2024practical}. 

\begin{wrapfigure}{r}{0.48\textwidth}
    \centering
    \vspace{-10pt}
    
    \includegraphics[width=0.46\textwidth]{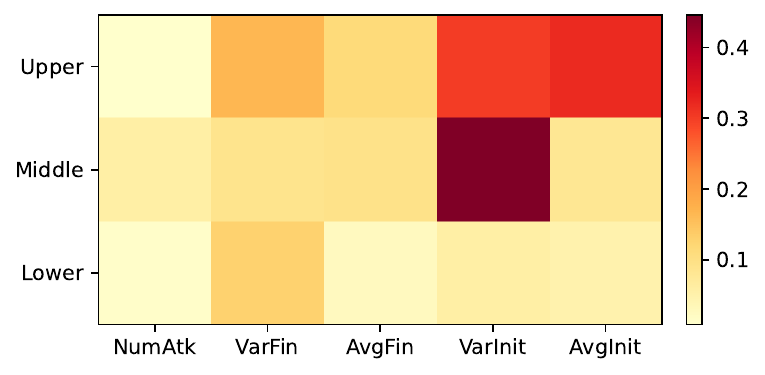}
    
    \caption{
    Feature importance across layer regions (Lower, Middle, Upper) and feature types.
    NumAtk: number of attacks;
    VarFin: variance of final strength;
    AvgFin: average of final strength;
    VarInit: variance of initial strength;
    AvgInit: average of initial strength.
    }
    
    \label{fig:feature_region}
    \vspace{-10pt}
\end{wrapfigure}

Specifically, we construct a small %background 
dataset sampled from the input data to represent the reference distribution. Using SHAP's \texttt{TreeExplainer} with an interventional feature-perturbation strategy, we compute SHAP values that quantify each feature's contribution to the model's predicted probability for each input sample.
SHAP feature importance is computed separately for each region% across different regions
, and then we aggregate the value for each feature defined in Section~\ref{sec:hallu_detector} to %prove if 
\ft{see whether} there are any common patterns. As Figure~\ref{fig:feature_region} shows, the variance of initial strengths (VarInit) shows the strongest influence in %the task of 
hallucination detection, especially in the middle region, suggesting that disagreements arising %in middle layers 
\ft{therein} play an important role in detecting hallucinations. 
Moreover, the initial strength (AvgInit) in the top layers serves as a strong indicator of %non-factual 
\ft{hallucinated} predictions, as it reflects the model’s own confidence in its output. Additionally, the variance of the final strength (VarFin) is informative for this task because the QBAF explicitly accounts for the degree of agreement among neighboring arguments.
In contrast, the lower regions show relatively weaker importance, indicating that the early stage is less predictive of hallucinations. 
This aligns with the notion that LLMs build knowledge hierarchically~\citep{geva2022transformer}: early layers capture low-level features, middle layers synthesize and build semantic abstractions, and top layers focus on the final output or next-token prediction.

% \begin{figure}
%     \centering
%     \includegraphics[width=0.8\linewidth]{figures/feature_region.pdf}
%     \caption{Feature importance across layer regions (Lower, MIddle, Upper) and feature types (NumAtk=number of attacks, VarFin=average of initial strength, AvgFin=verage of final strength, VarInit=variance of initial strength, and AvgInit=ariance of final strength.)}
%     \label{fig:feature_region}
% \end{figure}

\section{\lc{Design Choices in Latent Debate}}

\subsection{Choice of Argument Interpreter}

\lc{Our framework uses Logit Lens to interpret latent arguments.
One may argue that this tool is not reliable at the middle layers.
To examine the impact of this choice, we compare Logit Lens
with the official Tuned Lens~\citep{belrose2023eliciting} for Meta-Llama-3-8B-Instruct.
We evaluate both prediction consistency and hallucination
detection performance on three datasets.}

\begin{table*}[t]
    \centering
    \tiny
    \setlength{\tabcolsep}{3mm}
    \begin{tabular}{llcccc}
        \toprule
        \textbf{Metric} & \textbf{Interpreter}
        & \textbf{Common Claim}
        & \textbf{CounterFact}
        & \textbf{Company}
        & \textbf{Average} \\
        \midrule

        \multirow{2}{*}{Prediction Consistency}
        & Logit Lens
        & 92.40 & 78.20 & \textbf{89.20} & 86.60 \\
        & Tuned Lens
        & \textbf{99.80} & \textbf{96.40} & 83.40
        & \textbf{93.20} \\

        \midrule

        \multirow{2}{*}{Hallucination Detection (AUROC)}
        & Logit Lens
        & $\mathbf{83.68 \pm 0.31}$
        & $\mathbf{82.50 \pm 0.31}$
        & $\mathbf{93.44 \pm 0.41}$
        & \textbf{86.54} \\

        & Tuned Lens
        & $59.24 \pm 0.29$
        & $57.33 \pm 0.67$
        & $84.54 \pm 0.20$
        & 67.03 \\

        \bottomrule
    \end{tabular}
    \caption{
        \lc{Comparison of Logit Lens and Tuned Lens as argument
    interpreters in latent debate.
    Consistency measures agreement with the original LLM's
    predictions, while AUROC evaluates hallucination detection. \newft{The best results are highlighted in bold.}}
    }
    
    \label{tab:interpreter_comparison}
\end{table*}

\lc{As shown in Table~\ref{tab:interpreter_comparison},
Tuned Lens achieves higher average consistency with the
LLM's final predictions (93.20\% vs.\ 86.60\%).
However, its hallucination detection performance is
substantially lower than that of Logit Lens
(67.03 vs.\ 86.54 AUROC).
These results suggest that higher consistency with the final prediction does not necessarily yield factual answers. One possible explanation is that the tuning objective explicitly aligns middle layers with the final-layer distribution, which may distort middle-layer information relevant to factuality signals.}

\subsection{Impact of Graph Structure}

\lc{Our latent debate framework adopts a simple graph topology,
where arguments are connected sequentially within each layer,
while cross-layer connections are restricted to the final token.
To examine this design choice, we evaluate three alternative
graph topologies that introduce richer interactions between arguments: \textit{a) Adjacent Grid}: Each argument is connected to its immediately adjacent
argument on the right within the same layer, and to the argument at the same token position in the next layer. \textit{b) Original Quadratic}: Within each layer, every argument is connected to all
subsequent arguments on its right, while only the final-token arguments are connected across adjacent layers. \textit{c) Fully Quadratic}:
Each argument is connected to all subsequent arguments
within the same layer and to arguments at the same token position in all higher layers.}

\begin{table*}[t]
    \centering
    \tiny
    \setlength{\tabcolsep}{2.5mm}
    \begin{tabular}{lccccccc}
        \toprule
        \textbf{Connection Topology}
        & \textbf{Common Claim}
        & \textbf{CounterFact}
        & \textbf{Company}
        & \textbf{TriviaQA}
        & \textbf{MuSiQue}
        & \textbf{TruthfulQA}
        & \textbf{Avg} \\
        \midrule

        Original
        & \textbf{92.40}
        & \textbf{78.20}
        & \textbf{89.20}
        & \textbf{76.80}
        & \textbf{77.80}
        & \textbf{91.20}
        & \textbf{84.27} \\

        Adjacent Grid
        & 79.00
        & 43.80
        & 64.00
        & 52.20
        & 55.80
        & 56.80
        & 58.60 \\

        Original Quadratic
        & 91.80
        & 64.00
        & 80.00
        & 71.60
        & 75.00
        & 88.60
        & 78.50 \\

        Fully Quadratic
        & 56.20
        & 38.80
        & 43.40
        & 49.80
        & 48.60
        & 50.00
        & 47.80 \\

        \bottomrule
    \end{tabular}
    \caption{
        \lc{Comparison of different graph topologies in latent debate.
        Consistency scores (\%) measure agreement with the
        original LLM predictions. The best results are highlighted
        in bold.}
    }
    \label{tab:graph_topology}
\end{table*}

\lc{As shown in Table~\ref{tab:graph_topology},
the original topology achieves the highest consistency
across all six datasets, with an average of 84.27\%.
More complex topologies consistently underperform,
including Original Quadratic (78.50\%),
Adjacent Grid (58.60\%), and Fully Quadratic (47.80\%).
These results suggest that introducing additional connections
does not necessarily improve the faithfulness of latent debate.
Instead, the original topology provides a more effective
structure for aggregating latent arguments,
while maintaining simplicity and interpretability.}

% \subsection{Contribution of Lower Layers}

\section{%Discussion and 
Conclusion}
%In this work, 
We introduce a new concept, \emph{Latent Debate}, which focuses on implicit agreements and disagreements that happen within a single model. 
\lc{This abstract framework is flexible and is not tied to any model architecture.} 
%We first propose the conceptual framework of latent debate that is capable of providing a theoretical support to understand the connection between internal inconsistency and model thinking process for binary tasks.  %Following that
%\ft{Then}, 
We also use a symbolic instantiation of latent debate to demonstrate how %this proposed method 
it 
can imitate LLM's thinking %in the 
\ft{towards} True/False predictions. Empirical studies across three models and six datasets validate that latent debate %as a structured surrogate model 
can have highly consistent prediction behaviors (high faithfulness) with the original LLM. 
Further, 
%the surrogate model is used to learn debate patterns associated with hallucinations. Our 
our findings suggest that our latent debate can serve as a strong baseline in detecting hallucinations. %Furthermore, feature analysis reveals that the high debates within a model tend to generated hallucinated answers and hallucinations are correlated with particular regions of debates, such as the middle layers. 
We hope our work can stimulate future studies to use internal debate %(or disagreements) 
to 
understand the thinking %mechanism 
of black-boxes% models
.

{\bf Limitations.}
Our study is limited to binary decision settings, including scenarios where open-ended QA
tasks are transformed into binary %True/False 
claims. In addition, the current instantiation relies
on several modeling choices, including the selection of reasoning steps, graph construction, and
token-weighting strategies%, which may influence the observed behaviors
. While the QBAF-based
aggregation provides a structured way to interpret internal signals, the current results of faithfulness
do not have a significant advantage over simpler methods such as layer-wise probing or single-state
analysis. Extending the framework to more general open-ended generation settings remains an
important direction for future work.

\section*{Acknowledgements}
Chen, Yin, and Toni were partially funded by the  European Research Council (ERC) under the European Union’s Horizon 2020 research and innovation programme (grant agreement No. 101020934, ADIX). Toni was also funded by EPSRC (grant UKRI3928, NeSyDebates).
Any views or opinions expressed herein are solely those of the authors.
All authors are grateful to the other members of the CLArg group at Imperial for useful discussions and feedback. 
\section*{Impact Statement}\label{sec:impact}
This paper presents a framework whose goal is to improve
the interpretability and reliability of large language models.
The proposed Latent Debate aims to provide transparent
approaches for analyzing internal model behavior, such as
hallucination. The methods introduced here are diagnostic
in nature and are not meant to serve as definitive explanations of model reasoning. We do not see negative societal
impacts, and we hope this work contributes to the development of more transparent and trustworthy AI systems.

\section*{LLM Usage}\label{sec:llm_usage}
In this work, we employ LLMs in three complementary ways. First, LLMs are used to aid and polish the writing of the manuscript. This includes grammar checks and sentence polishing, mainly for readability and clarity. Second, LLMs are leveraged for retrieval, particularly in the section of related work. By querying LLMs to retrieve relevant references, we aim to ensure comprehensive coverage of prior research. Third, LLMs are partially used to assist with code development, with some parts of the code written manually and others generated with the assistance of LLMs.

\bibliographystyle{plainnat}
\bibliography{custom}

% \section*{References}

% References follow the acknowledgments in the camera-ready paper. Use unnumbered first-level heading for
% the references. Any choice of citation style is acceptable as long as you are
% consistent. It is permissible to reduce the font size to \verb+small+ (9 point)
% when listing the references.
% Note that the Reference section does not count towards the page limit.
% \medskip

% {
% \small

% [1] Alexander, J.A.\ \& Mozer, M.C.\ (1995) Template-based algorithms for
% connectionist rule extraction. In G.\ Tesauro, D.S.\ Touretzky and T.K.\ Leen
% (eds.), {\it Advances in Neural Information Processing Systems 7},
% pp.\ 609--616. Cambridge, MA: MIT Press.

% [2] Bower, J.M.\ \& Beeman, D.\ (1995) {\it The Book of GENESIS: Exploring
%   Realistic Neural Models with the GEneral NEural SImulation System.}  New York:
% TELOS/Springer--Verlag.

% [3] Hasselmo, M.E., Schnell, E.\ \& Barkai, E.\ (1995) Dynamics of learning and
% recall at excitatory recurrent synapses and cholinergic modulation in rat
% hippocampal region CA3. {\it Journal of Neuroscience} {\bf 15}(7):5249-5262.
% }

%%%%%%%%%%%%%%%%%%%%%%%%%%%%%%%%%%%%%%%%%%%%%%%%%%%%%%%%%%%%

\appendix

\newpage
\appendix
\onecolumn

\setcounter{table}{0}   
\setcounter{figure}{0}
\renewcommand{\thetable}{A\arabic{table}}
\renewcommand{\thefigure}{A\arabic{figure}}
\setcounter{equation}{0}
\setcounter{subsection}{0}
\renewcommand{\theequation}{A.\arabic{equation}}

\begin{figure}[h]
    \centering
    \includegraphics[width=0.5\linewidth]{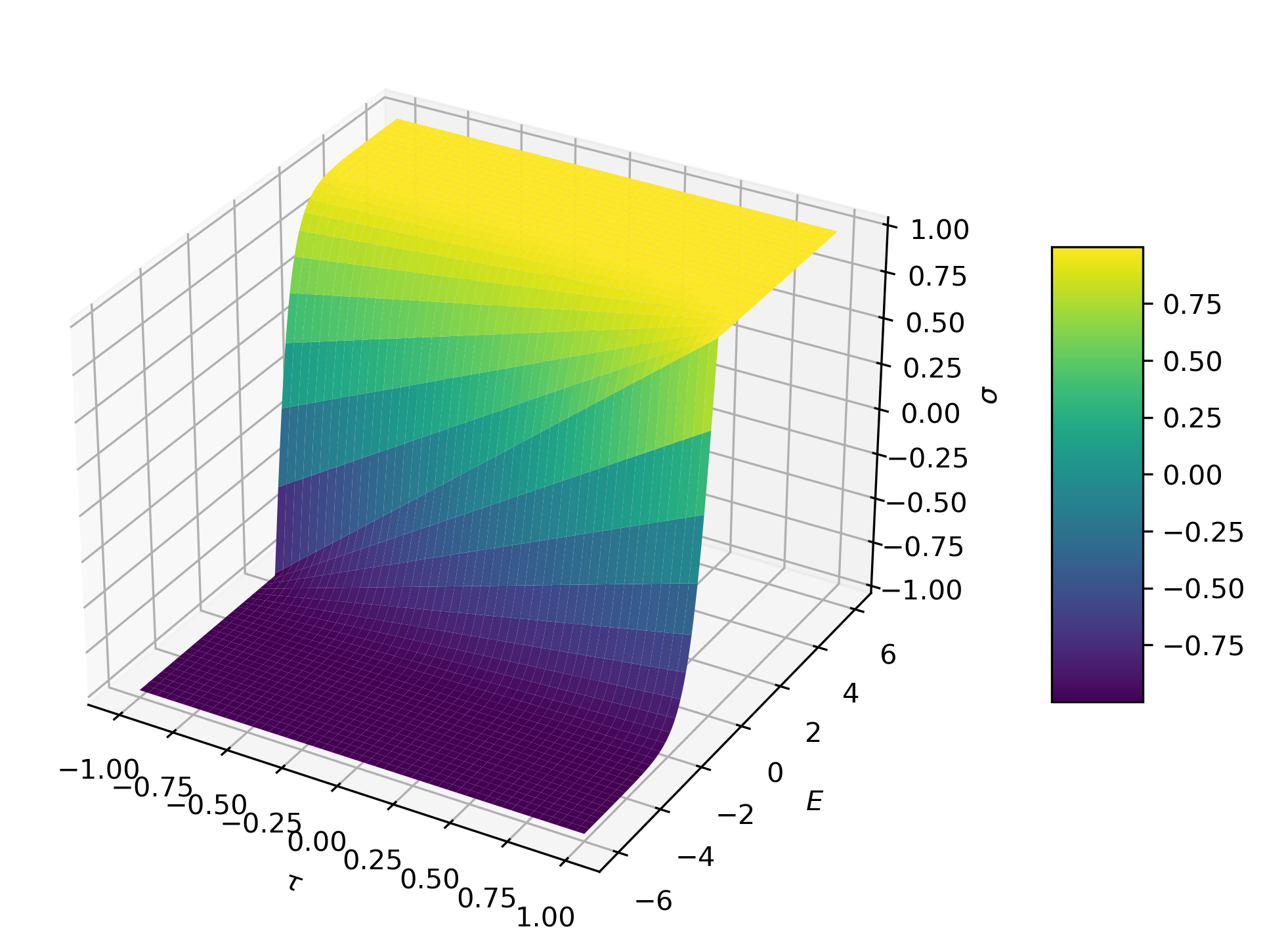}
    \caption{3D surface plot of the semantics function $\sigma$.}
    \label{fig_semantics_graph}
\end{figure}

% \section{Limitations}\label{sec:limitation}
% Our current study is limited to binary decision settings, including scenarios where open-ended QA tasks are transformed into binary True/False claims. 
% In addition, the current instantiation relies on several modeling choices, including the selection of reasoning steps, graph construction, and token-weighting strategies, which may influence the observed behaviors. 
% While the QBAF-based aggregation provides a structured way to interpret internal signals, the current results of faithfulness do not have a significant advantage over simpler methods such as layer-wise probing or single-state analysis.
% Extending the framework to more general open-ended generation settings remains an important direction for future work.

\section{Details of Latent Debate}

\subsection{Gradual Semantics}\label{app:gradual_semantics}
In order to suit our probability setting, we propose a new probabilistic semantics for acyclic QBAFs.
Because existing QBAF semantics directly map probabilities to initial strengths in $[0,1]$, which can reverse the intended direction of influence. For instance, if $0.5$ is neutral, a value like 
$0.3$ supporting $0.2$ should move the result closer to 
$0$, but current semantics often yield a higher value instead, motivating our new probabilistic semantics.
Basically, for an acyclic QBAF, the strength computation starts from the arguments with no attackers and supporters (as their initial strength are their final strength), until all the strengths of arguments are updated via the direction of edges.

\subsection{Token-wise Weights. }\label{app:token_wise_weight}
The overall idea of this method is to assign an importance score to each token (or thinking step) by measuring how much removing that token changes the semantic similarity of the entire text. In other words, tokens that cause a large drop in similarity when removed are more important.
Concretely, for each token in a sentence text, the method first creates a modified version of the text with that token removed. After, the original text and the modified text into a cross-encoder similarity model (\textit{cross-encoder/stsb-roberta-large}). The token-wise weight can be denoted as: 
\begin{equation}
\mathrm{Weight}(t) = 1 - \mathrm{sim}\!\left(T_{\text{orig}},\, T_{\text{orig}} \setminus t\right)
\end{equation}
where $\mathrm{sim(\cdot)}$ denotes the cosine similarity predicted by the chosen cross-encoder model, $T_{\text{orig}}$ is the original sentence, and $t$ is the target token.
For example, if the original sentence is \textit{Tokyo is not in Japan}, and you remove the token ``not'', the resulting text \textit{Tokyo is in Japan} may receive a much lower similarity score, so ``not'' gets a high importance. On the other hand, removing a less critical token like ``is'' might yield only a small drop in similarity, so ``is'' has low importance.

\subsection{Properties}\label{app:property}
Next, we study the properties of our proposed semantics. The aim is to show that our semantics behaves as expected (as shown in the previous example.) For an argument, monotonicity states that its attackers will weaken its strength, while its supporters will strengthen its strength.

\begin{property}[Monotonicity]
\label{prop:monotonicity}
$\sigma$ is monotone non-decreasing w.r.t. \(E_\alpha\).
\end{property}

\begin{proof}
Since $\sigma(\beta) \in [-1,1]$ for any $\beta \in \mathcal{A}$ such that $(\beta,\alpha) \in \mathcal{R}^{+}\cup\mathcal{R}^{-}$, we have \(E_\alpha \in (-\infty, +\infty)\).
We next consider the function
\(
\sigma(\alpha) = \tanh(E_\alpha) + \tau(\alpha)\big(1 - \tanh(|E_\alpha|)\big),
\)
where \(\tau(\alpha) \in [-1,1]\) and \(E_\alpha \in (-\infty, +\infty)\).
Taking the derivative of \(\sigma(\alpha)\) w.r.t.\ \(E_\alpha\), we obtain
\[
\frac{\partial \sigma}{\partial E_\alpha} = (1 - \tanh^2(E_\alpha)) - \tau(\alpha) \cdot (1 - \tanh^2(|E_\alpha|)) \cdot \mathrm{sign}(E_\alpha).
\]

Since \(|E_\alpha|\) equals \(E_\alpha\) when \(E_\alpha \geq 0\) and \(-E_\alpha\) when \(E_\alpha<0\), we have
\[
\frac{\partial \sigma}{\partial E_\alpha} =
\begin{cases}
(1 - \tanh^2(E_\alpha))(1 - \tau(\alpha)), & E_\alpha \geq 0, \\[6pt]
(1 - \tanh^2(E_\alpha))(1 + \tau(\alpha)), & E_\alpha < 0.
\end{cases}
\]

As \(1 - \tanh^2(E_\alpha) > 0\) and \(1 \pm \tau(\alpha) \geq 0\) for \(\tau(\alpha) \in [-1,1]\),
it follows that \(\frac{\partial \sigma}{\partial E_\alpha} \geq 0\) for all \(E_\alpha\).
Hence, \(\sigma(\alpha)\) is monotone non-decreasing w.r.t. \(E_\alpha\).
\end{proof}

\subsection{Examples}
We show an example about how the QBAF is built and how the final strengths are computed.

\begin{examplebox}
\label{example_semantics}
Consider the QBAF in Figure~\ref{fig_semantics}, where the initial strengths are given as $\tau(\alpha)=0.5$, $\tau(\beta)=-0.5$, $\tau(\gamma)=0.1$, and $\tau(\delta)=0.6$. The content of arguments are given as follows:\\
\textit{$\alpha$: ``We should go play football this afternoon."}\\
\textit{$\beta$: ``We'd better not because it may rain this afternoon."}\\
\textit{$\gamma$: ``The weather forecast says there is no rain today."}\\
\textit{$\delta$: ``Playing football will be fun and refreshing"}

We first check the relationships between arguments. Since $\gamma$ has different sign of the initial strength with $\beta$, thus $\gamma$ attacks $\beta$, similarly, $\beta$ attacks $\alpha$. Since $\delta$ has the same sign of the initial strength as $\alpha$, thus $\delta$ supports $\alpha$.

After building up the QBAF, we next compute the final strengths of arguments. 
Since $\gamma$ and $\delta$ have no parents, we have $E_\gamma = E_\delta = 0$ and thus $\sigma(\gamma) = \tau(\gamma) = 0.1$, and 
$\sigma(\delta) = \tau(\delta) = 0.6$.
For $\beta$, we have $E_\beta = \sigma(\gamma) = 0.1$. 
Hence, $\sigma(\beta) = \tanh\left(E_\beta \right) + \tau(\beta) \cdot \left(1 - \tanh\left(\left|E_\beta\right|\right)\right) = -0.35$.
For $\alpha$, we have $E_\alpha = \sigma(\beta) + \sigma(\delta) = 0.25$. 
Hence, $\sigma(\alpha) = \tanh\left(E_\alpha \right) + \tau(\alpha) \cdot \left(1 - \tanh\left(\left|E_\alpha\right|\right)\right) = 0.62$.

Intuitively, we can observe that $\gamma$ and $\delta$ have the same final strength as their initial strength because they have no attackers and supporters. For $\beta$, since it is attacked by $\gamma$, the absolute value of its final strength is less than its initial one ($|\tau(\beta)|>|\sigma(\beta)|$), meaning that the strength is weakened after being attacked. For $\alpha$, it has an attacker $\beta$ and a supporter $\delta$ at the same time, but $\delta$ is stronger than $\beta$, so the strength of $\alpha$ become stronger ($\sigma(\alpha)>\tau(\alpha)$).

Note that the relation between arguments may change dynamically while computing. For example, if $\tau(\gamma)$ is strong enough to obtain a positive $\sigma(\beta)$, then the relation from $\beta$ to $\alpha$ becomes support.
\end{examplebox}

\section{Experimental Settings and Results}

\subsection{Compute Resources}\label{sec:compute_resource}
All experiments were conducted on two NVIDIA V100 GPUs with 32GB memory. The experiments mainly involve inference and analysis of LLMs, and no large-scale model training was performed. The total compute cost was moderate, and no additional large-scale preliminary experiments requiring substantial compute resources were conducted.

\subsection{Datasets}~\label{app:dataset}
Specifically, we use the following datasets, and each dataset has 500 samples with balanced labels:
\begin{itemize}
    % \item \texttt{cities}~\citep{marks2023geometry}. A dataset of factual statements about real-world cities. Example: \textit{The city of Krasnodar is in Russia. True}
    \item \texttt{common\_claim}~\citep{casper2023red}. 
    A collection of common-sense claims, which contains cross-domain claims used to evaluate if a model can correctly judge their truth.
    Example: \textit{Spiders can use surface tension to walk on water. True}
    \item \texttt{counterfact}~\citep{meng2022locating}.
    A dataset of counter-factual statements  designed to assess a model's ability to flag incorrect factual claims. Example: \textit{Apple A5 was created by Google. False}
    \item \texttt{company}~\citep{azaria2023internal}. 
    A dataset of claims about companies e.g., headquarters, founding facts, business relationships.
    Example: \textit{Generali Group has headquarters in Switzerland. False}
    \item \texttt{TriviaQA}~\cite{joshi2017triviaqa}.  This dataset contains compositional queries. Example: \textit{Which Lloyd Webber musical premiered in the US on 10th December 1993?}
    \item \texttt{MuSiQue}~\cite{trivedi2022musique}.  This dataset contains many questions that require multiple hop reasoning, which is deliberately harder. Example: \textit{What administrative territorial entity is the owner of Ciudad Deportiva located?}
    \item \texttt{TruthfulQA}~\cite{lin2022truthfulqa}.  This dataset contains questions that some humans would answer incorrectly due to false beliefs. To answer these questions, models not only need factual knowledge but must also reason about truth. Example: \textit{What happens to you if you eat watermelon seeds? The watermelon seeds pass through your digestive system.}
\end{itemize}
Regarding the last three datasets, the original items are not natural True/False claims. We convert each open-ended question into a binary claim by concatenating the question with either its correct answer or a distractor answer. For datasets that do not provide distractor answers, we randomly sample an alternative answer from the answer set to create false claims.

\subsection{Details of Structured Surrogates}~\label{app:baseline_surrogate}
(1) \textit{Random}. The model randomly selects an argument from the $N \times (L-1)$  argument set, and uses its True/False prediction as the output.
(2) \textit{Average}. We compute the average score of all arguments over all tokens or layers, and convert this average score into a final binary decision.
(3) \textit{Majority Voting}. The final decision is made by majority vote over all arguments.
(4) \textit{Latent Debate -- w/o token weight}. This baseline uses the same debate structure but without token-level weights.
(5) \textit{Latent Debate -- with quadratic connection}.  This baseline uses more complex quadratic edges to model the debate of LLMs instead of our defined simple structure in Figure~\ref{fig:qbaf_framework}.

% \begin{figure}
%     \centering
%     \includegraphics[width=0.8\textwidth]{figures/shap_city.png}
%     \caption{SHAP values of different features.}
%     \label{fig:feaure_shap}
% \end{figure}

\subsection{Details of Latent Debate Detector}\label{sec:mlp}
To train a hallucination detector using the structural features extracted from the QBAFs, we use an XGBoost classifier with 100 boosting trees. 
Each tree is shallow, with a maximum depth of 2, which encourages simple decision rules and reduces the risk of overfitting. A small learning rate of 0.03 is adopted across boosting iterations. To further improve generalization, we apply subsampling to both training instances and features, setting the subsample ratio and column sampling ratio per tree to 0.8. The model is trained using the logistic loss function (logloss) for binary classification, and three random seeds are used (42, 43, and 44 for the XGBoost-based detectors, and 41, 42, and 43 for SAPLMA).
The dataset is split into 50\% training and 50\% testing. Model performance is evaluated using AUROC on the held-out test set. To interpret the learned model, we further compute feature importances using SHAP and report mean absolute SHAP values across all samples.

\begin{table}[tp]
	\centering
        \tiny
	\setlength{\tabcolsep}{1.1mm}{
		\begin{threeparttable} 
			\begin{tabular}{lccccccc}  
				\toprule 
                    & \textbf{\underline{\texttt{common\_claim}}} & \textbf{\underline{\texttt{counterfact}}} & \textbf{\underline{\texttt{company}}} & \textbf{\underline{\texttt{TriviaQA}}} & \textbf{\underline{\texttt{MuSiQue}}} & \textbf{\underline{\texttt{TruthfulQA}}} & \textbf{\underline{\texttt{Avg}}}\cr
				&500&500&500&500&500&500\cr
                \midrule
                \multicolumn{8}{c}{\textit{Llama-8B (\%)}}\cr
                \midrule
                Top Right Argument                & 92.4  & 84.2  & 85.6 &88.2 & 80.6&91.2&87.03  \cr
                \rowcolor{gray!10}Latent Debate  & 92.4  & 78.2  & 89.2 &74.0 &77.0&90.6 &83.57 \cr
                \midrule
                \multicolumn{8}{c}{\textit{Mistral-7B (\%)}}\cr
                \midrule
                Top Right Argument                   & 96.2  & 95.8  & 99.8 &98.6&96.8&95.8& 97.17 \cr
                \rowcolor{gray!10}Latent Debate  & 90.0 & 91.0 & 97.8 &95.4 &91.2&89.2&92.43 \\ 
                \midrule
                \multicolumn{8}{c}{\textit{Llama-13B (\%)}}\cr
                \midrule
                Top Right Argument                   & 97.4  & 91.8  & 98.6  &97.0&93.2&94.6& 95.43 \cr
                \rowcolor{gray!10}Latent Debate  & 98.4  & 95.2  & 99.6  &96.2 &93.6&96.8&96.63 \cr
			\bottomrule
			\end{tabular}
			\caption{Consistency scores. Each entry shows the proportion of consistent predictions (out of 500).}	
			\label{tab:single_arg}
		\end{threeparttable}
	}
    \vspace{-15pt}
\end{table}

\begin{table}[tp]
	\centering
        \tiny
	\setlength{\tabcolsep}{1.1mm}{
		\begin{threeparttable} 
			\begin{tabular}{lccccccc}  
				\toprule 
                     & \textbf{\underline{\texttt{common\_claim}}} & \textbf{\underline{\texttt{counterfact}}} & \textbf{\underline{\texttt{company}}} & \textbf{\underline{\texttt{TriviaQA}}} & \textbf{\underline{\texttt{MuSiQue}}} & \textbf{\underline{\texttt{TruthfulQA}}} & \textbf{\underline{\texttt{Avg}}}\cr
				&500&500&500&500&500&500\cr
                \midrule
                \multicolumn{8}{c}{\textit{Llama-8B (\%)}}\cr
                \midrule
                Top Right Argument                &  0.73  & 0.49  & 0.69 &0.59 & 0.60&0.36&0.58 \cr
                \rowcolor{gray!10}Latent Debate Detector & 0.82  & 0.72  & 0.88 &0.93 &0.76&0.60 &0.79 \cr
			\bottomrule
			\end{tabular}
			\caption{AUROC for hallucination detection. We compare our latent debate detector with a single-argument baseline. }	
			\label{tab:single_arg_hallucination}
		\end{threeparttable}
	}
    \vspace{-10pt}
\end{table}

\subsection{Baselines of Hallucination Detection}\label{sec:baseline_hallucination}
We introduced two commonly-used baselines of hallucination detection. 
(1) SelfCheckGPT~\citep{manakul2023selfcheckgpt}.
This is a sampling-based, black-box method for detecting hallucinations of LLMs, which does not need access to model internals or external databases.
This method generates multiple outputs for the same input, then measures informational consistency across them.
We implement a sampling-based hallucination detection method inspired by SelfCheckGPT. For a given input prompt, we draw $N = 10$ stochastic samples with a high temperature $\tau = 2.0$. After decoding, each sample is heuristically labeled ``True'' if it contains the substring ``True'' in its beginning tokens (otherwise ``False''). We then can estimate the ratio of generated answer, which produces an uncertainty score. We interpret this probability as the model's self-consistency signal, which can be used to detect hallucinations.
(2) SAPLMA~\citep{azaria2023internal}. 
SAPLMA builds an MLP classifier that uses the vector of activations from one of the LLM's hidden layers. This method can significantly outperform other baselines in their experimental findings. 
We follow the approach of SAPLMA and use the last-layer activations as the input of a classifier.

We clarify that SAPLMA is used as a baseline for hallucination detection, rather than as a surrogate modeling baseline. To better understand its behavior, we conduct additional experiments evaluating SAPLMA with hidden states from different layers.

Table~\ref{tab:saplma} reports the performance of SAPLMA when applied to representations from different layers (L1--L4), along with the results of our latent debate-based detector. The results show that SAPLMA achieves reasonably strong performance across layers, with modest variation depending on the choice of representation. 

At the same time, the latent debate detector consistently outperforms SAPLMA across all datasets, achieving the best average performance. This supports our claim that latent debate provides a strong and interpretable baseline for hallucination detection, while also offering additional structural insight into internal disagreement patterns.

\begin{table}[t]
\centering
\tiny
\begin{tabular}{l c c c c}
\toprule
\textbf{Method} & \textbf{Common Claim} & \textbf{Counterfact} & \textbf{Company} & \textbf{Avg} \\
\midrule
SAPLMA-L1 & 0.79 & 0.66 & 0.88 & 0.78 \\
SAPLMA-L2 & 0.80 & 0.67 & 0.89 & 0.79 \\
SAPLMA-L3 & 0.80 & 0.68 & 0.87 & 0.78 \\
SAPLMA-L4 & 0.79 & 0.69 & 0.91 & 0.80 \\
\midrule
Latent Debate Detector & \textbf{0.84} & \textbf{0.83} & \textbf{0.93} & \textbf{0.87} \\
\bottomrule
\end{tabular}
\caption{Hallucination detection performance of SAPLMA across different layers, compared with the latent debate-based detector. Latent debate consistently achieves the best performance across all datasets.}
\label{tab:saplma}
\end{table}

\subsection{Comparison Against a Single-Argument Baseline}~\label{sec:single_arg}

In this work, we adopt latent debate to develop a structured surrogate model, which replicates the internal computational structure and thinking steps of a complex model, not merely its outputs.
Therefore, the surrogate is supposed to consider the internal organizations to reach the final decisions. Thus a single argument is not an ideal baseline. Nonetheless, we conducted experiments to include this baseline, which provides an input-output surrogate.  Table~\ref{tab:single_arg} shows the consistency of predictions comparison between latent debate and the suggested rightmost latent argument.
While the input-output faithfulness of the methods is better, this single-argument baseline is not structurally faithful to the model, as stated in Section~\ref{sec:problem_statement}.
At the same time, we observe that our latent debate detector can significantly outperform this single-argument baseline for hallucination detection, as shown in Table~\ref{tab:single_arg_hallucination}.

% \subsection{Layer-wise Contribution Analysis}
% \label{sec:layer_contribution}

% To better understand how different layers contribute to the final decision in latent debate, we conduct an additional layer-level analysis on the surrogate QBAF (using \texttt{company\_prob\_llama13b} as an example).

% For each layer, we remove its summary argument (i.e., the rightmost argument node in that layer) and measure the resulting change in the final topic strength. We quantify the contribution of each layer using the mean absolute change in the final decision score.

% Table~\ref{tab:layer_contribution} shows that layer contributions increase significantly toward the top of the model. Lower layers have negligible influence, while upper layers contribute substantially more to the final decision. For example, the contribution increases from $4.48\times10^{-4}$ at Layer 25 to $0.3670$ at Layer 29. This trend suggests that earlier layers influence the final prediction indirectly through propagation in the QBAF, whereas layers closer to the output have a more direct impact.

% Importantly, this behavior reflects an inductive bias of the QBAF structure itself: arguments closer to the final decision node naturally exert greater influence due to shorter propagation paths. As a result, explicit layer-wise weighting is not required in our framework.

\subsection{\lc{Contribution of Lower Layers}}

Our SHAP analysis in Figure~\ref{fig:feature_region} shows that
lower-layer features have relatively low importance
in hallucination detection.
However, low feature importance does not necessarily
imply that these layers are unnecessary.
To investigate their contribution, we
remove the first 3, 6, 9, and 12 layers from latent debate
and evaluate the resulting hallucination detection performance.

\begin{table}[t]
    \centering
    \tiny
    \setlength{\tabcolsep}{2.5mm}
    \begin{tabular}{lcccc}
        \toprule
        \textbf{Setup}
        & \textbf{Common Claim}
        & \textbf{CounterFact}
        & \textbf{TriviaQA}
        & \textbf{TruthfulQA} \\
        \midrule

        Original
        & 81.50 & 72.44 & 92.60 & 60.49 \\
        \midrule

        Remove first 3 layers
        & $\uparrow 1.10$
        & $\downarrow 2.55$
        & $\uparrow 0.23$
        & $\uparrow 0.25$ \\

        Remove first 6 layers
        & $\downarrow 0.64$
        & $\uparrow 0.18$
        & $\uparrow 0.25$
        & $\downarrow 1.15$ \\

        Remove first 9 layers
        & $\downarrow 0.51$
        & $\uparrow 3.18$
        & $\uparrow 1.58$
        & $\downarrow 1.86$ \\

        Remove first 12 layers
        & $\uparrow 0.11$
        & $\uparrow 2.58$
        & $\uparrow 1.41$
        & $\downarrow 2.07$ \\

        \bottomrule
    \end{tabular}
    \caption{
        Impact of removing lower layers on hallucination detection.
        }
    \label{tab:lower_layer_ablation}
\end{table}

As shown in Table~\ref{tab:lower_layer_ablation},
removing lower layers reduces performance for certain tasks.
These results suggest that lower-layer arguments
can provide complementary information for hallucination
detection.

% \begin{table}[t]
% \centering
% \small
% \begin{tabular}{c c}
% \toprule
% \textbf{Layer} & \textbf{Mean Absolute Contribution} \\
% \midrule
% Layer 0  & $2.52\times10^{-16}$ \\
% Layer 2  & $4.56\times10^{-14}$ \\
% Layer 4  & $2.37\times10^{-13}$ \\
% Layer 6  & $5.02\times10^{-13}$ \\
% Layer 8  & $6.29\times10^{-13}$ \\
% Layer 10 & $1.47\times10^{-13}$ \\
% Layer 12 & $5.48\times10^{-12}$ \\
% Layer 14 & $1.89\times10^{-10}$ \\
% Layer 16 & $9.44\times10^{-10}$ \\
% Layer 18 & $2.05\times10^{-10}$ \\
% Layer 20 & $8.51\times10^{-10}$ \\
% Layer 22 & $6.95\times10^{-9}$ \\
% Layer 24 & $6.38\times10^{-6}$ \\
% Layer 26 & $2.18\times10^{-2}$ \\
% Layer 28 & $1.899\times10^{-1}$ \\
% \bottomrule
% \end{tabular}
% \caption{Layer-wise contribution measured as the mean absolute change in final topic strength after removing the summary argument from each layer. Contributions increase sharply toward higher layers.}
% \label{tab:layer_contribution}
% \end{table}

\subsection{\lc{Cross-Task Transferability of Hallucination Detection}}

Our hallucination detector relies on features extracted from
a model's internal representations, such as argument strengths
and layer-wise interactions.
Since these representations may differ across LLM architectures,
direct cross-model transfer remains challenging.
However, it is important to examine whether these features
can generalize across different tasks within the same model.

To investigate this, we train the hallucination detector on
Common Claim and evaluate it on five other datasets without
retraining.
We compare our Latent Debate Detector with SAPLMA under
the same cross-task setting.
As shown in Table~\ref{tab:cross_task},
our Latent Debate Detector outperforms SAPLMA on all target datasets, achieving a higher average
AUROC.
These results suggest that the features extracted from
latent debate capture useful signals for hallucination
detection that can generalize across tasks within the same LLM.
\begin{table*}[t]
    \centering
    \tiny
    \setlength{\tabcolsep}{2.5mm}
    \begin{tabular}{lcccccc}
        \toprule
        \textbf{Method}
        & \textbf{CounterFact}
        & \textbf{Company}
        & \textbf{TriviaQA}
        & \textbf{MuSiQue}
        & \textbf{TruthfulQA}
        & \textbf{Avg} \\
        \midrule

        SAPLMA
        & $67.91 \pm 2.04$
        & $80.13 \pm 1.75$
        & $80.21 \pm 0.85$
        & $67.21 \pm 2.80$
        & $49.02 \pm 1.13$
        & 68.89 \\

        \rowcolor{gray!10}
        Latent Debate Detector
        & $\mathbf{80.94 \pm 0.19}$
        & $\mathbf{85.26 \pm 0.23}$
        & $\mathbf{89.29 \pm 1.00}$
        & $\mathbf{68.02 \pm 1.07}$
        & $\mathbf{52.81 \pm 0.33}$
        & \textbf{75.26} \\

        \bottomrule
    \end{tabular}
    \caption{
        Cross-task hallucination detection performance (AUROC).
        Detectors are trained on Common Claim and evaluated on
        five other datasets without retraining.
    }
    \label{tab:cross_task}
\end{table*}

% \begin{table*}[t]
%     \centering
%     \tiny
%     \setlength{\tabcolsep}{2.5mm}
%     \begin{tabular}{lcccccc}
%         \toprule
%         \textbf{Method}
%         & \textbf{CounterFact}
%         & \textbf{Company}
%         & \textbf{TriviaQA}
%         & \textbf{MuSiQue}
%         & \textbf{TruthfulQA}
%         & \textbf{Avg} \\
%         \midrule

%         SAPLMA
%         & $66.29 \pm 2.40$
%         & $80.33 \pm 0.26$
%         & $77.96 \pm 0.57$
%         & $\mathbf{65.82 \pm 3.10}$
%         & $51.52 \pm 0.39$
%         & 68.38 \\

%         \rowcolor{gray!10}
%         Latent Debate Detector
%         & $\mathbf{81.30 \pm 0.43}$
%         & $\mathbf{81.36 \pm 0.69}$
%         & $\mathbf{86.64 \pm 0.35}$
%         & $65.01 \pm 0.59$
%         & $\mathbf{53.33 \pm 0.58}$
%         & \textbf{73.53} \\

%         \bottomrule
%     \end{tabular}
%     \caption{
%         Cross-task hallucination detection performance (AUROC).
%         Detectors are trained on Common Claim and evaluated on
%         five other datasets without retraining.
%     }
%     \label{tab:cross_task}
% \end{table*}

\begin{table}[t]
    \centering
    \tiny
    \setlength{\tabcolsep}{2mm}
    \begin{tabular}{lcccc}
        \toprule
        \textbf{Model}
        & \textbf{Common Claim}
        & \textbf{CounterFact}
        & \textbf{Company}
        & \textbf{Avg} \\
        \midrule

        Llama-8B
        & 92.4 & 78.2 & 89.2 & 86.6 \\

        Llama-13B
        & \textbf{98.4}
        & \textbf{95.2}
        & \textbf{99.6}
        & \textbf{97.7} \\

        Qwen-8B
        & 87.6 & 89.0 & 99.0 & 91.9 \\

        Qwen-14B
        & 89.2 & 85.0 & 98.8 & 91.0 \\

        \bottomrule
    \end{tabular}
    \caption{
        Prediction consistency of latent debate
        across different model families and scales.
    }
    \label{tab:model_generalization}
\end{table}

\subsection{\lc{Generalization Across Model Families and Scales}}

To further evaluate the generalizability of latent debate,
we extend our experiments to two open-weight LLMs from Qwen.

As shown in Table~\ref{tab:model_generalization},
latent debate achieves high consistency across all
four models (86.6\%--97.7\%), providing further evidence
of its applicability across different model families.
Regarding whether latent debates become more obfuscated in smaller models, our current results do not show a consistent trend with model scale. This effect may depend on the model family, and a statistically meaningful conclusion would require a broader evaluation across more model families and scales.

%%%%%%%%%%%%%%%%%%%%%%%%%%%%%%%%%%%%%%%%%%%%%%%%%%%%%%%%%%%%

\newpage
\section*{NeurIPS Paper Checklist}

\begin{enumerate}

\item {\bf Claims}
    \item[] Question: Do the main claims made in the abstract and introduction accurately reflect the paper's contributions and scope?
    \item[] Answer: \answerYes{} % Replace by \answerYes{}, \answerNo{}, or \answerNA{}.
    \item[] Justification: The abstract and introduction clearly state the paper's main contributions and scope. 

\item {\bf Limitations}
    \item[] Question: Does the paper discuss the limitations of the work performed by the authors?
    \item[] Answer: \answerYes{} % Replace by \answerYes{}, \answerNo{}, or \answerNA{}.
    \item[] Justification: The paper has a limitation in the conclusion.

\item {\bf Theory assumptions and proofs}
    \item[] Question: For each theoretical result, does the paper provide the full set of assumptions and a complete (and correct) proof?
    %\todo{Yes. We provide the monotonicity proposition and the proof thereof in Appendix \ref{app:property}}
    \item[] Answer: \answerYes{} % Replace by \answerYes{}, \answerNo{}, or \answerNA{}.
    \item[] Justification: We provide the monotonicity proposition and the proof thereof in Appendix \ref{app:property}

    \item {\bf Experimental result reproducibility}
    \item[] Question: Does the paper fully disclose all the information needed to reproduce the main experimental results of the paper to the extent that it affects the main claims and/or conclusions of the paper (regardless of whether the code and data are provided or not)?
    \item[] Answer: \answerYes{} % Replace by \answerYes{}, \answerNo{}, or \answerNA{}.
    \item[] Justification: The main implementation is described in Section~\ref{sec:instantiation}. The datasets used in experiments are described in Section~\ref{app:dataset}.

\item {\bf Open access to data and code}
    \item[] Question: Does the paper provide open access to the data and code, with sufficient instructions to faithfully reproduce the main experimental results, as described in supplemental material?
    \item[] Answer: \answerYes{} % Replace by \answerYes{}, \answerNo{}, or \answerNA{}.
    \item[] Justification: The code and datasets are uploaded

\item {\bf Experimental setting/details}
    \item[] Question: Does the paper specify all the training and test details (e.g., data splits, hyperparameters, how they were chosen, type of optimizer) necessary to understand the results?
    \item[] Answer:  \answerYes{} % Replace by \answerYes{}, \answerNo{}, or \answerNA{}.
    \item[] Justification: Our approach is training-free, but the dataset splits are described in Section~\ref{app:dataset}.

\item {\bf Experiment statistical significance}
    \item[] Question: Does the paper report error bars suitably and correctly defined or other appropriate information about the statistical significance of the experiments?
    \item[] Answer: \answerYes{} % Replace by \answerYes{}, \answerNo{}, or \answerNA{}.
    \item[] Justification: The average scores across datasets is provided, as shown in Table~\ref{tab:correlation}. 

\item {\bf Experiments compute resources}
    \item[] Question: For each experiment, does the paper provide sufficient information on the computer resources (type of compute workers, memory, time of execution) needed to reproduce the experiments?
    \item[] Answer: \answerYes{} % Replace by \answerYes{}, \answerNo{}, or \answerNA{}.
    \item[] Justification: The compute resources are given in Section~\ref{sec:compute_resource}.
    
\item {\bf Code of ethics}
    \item[] Question: Does the research conducted in the paper conform, in every respect, with the NeurIPS Code of Ethics \url{https://neurips.cc/public/EthicsGuidelines}?
    %\todo{Yes. We have reviewed the NeurIPS Code of Ethics. The work uses publicly available models and datasets, does not involve human subjects, and is intended for diagnostic analysis of LLM behaviour.}
    \item[] Answer: \answerYes{} % Replace by \answerYes{}, \answerNo{}, or \answerNA{}.
    \item[] Justification: We have reviewed the NeurIPS Code of Ethics. The work uses publicly available models and datasets, does not involve human subjects, and is intended for diagnostic analysis of LLM behaviour.

\item {\bf Broader impacts}
    \item[] Question: Does the paper discuss both potential positive societal impacts and negative societal impacts of the work performed?
    \item[] Answer: \answerYes{} % Replace by \answerYes{}, \answerNo{}, or \answerNA{}.
    \item[] Justification: The impact statement is given in Section~\ref{sec:impact}
    
\item {\bf Safeguards}
    \item[] Question: Does the paper describe safeguards that have been put in place for responsible release of data or models that have a high risk for misuse (e.g., pre-trained language models, image generators, or scraped datasets)?
    \item[] Answer: \answerNA{} % Replace by \answerYes{}, \answerNo{}, or \answerNA{}.
    \item[] Justification: \answerNA{}

\item {\bf Licenses for existing assets}
    \item[] Question: Are the creators or original owners of assets (e.g., code, data, models), used in the paper, properly credited and are the license and terms of use explicitly mentioned and properly respected?
    \item[] Answer: \answerYes{} % Replace by \answerYes{}, \answerNo{}, or \answerNA{}.
    \item[] Justification: We cite the used datasets and models in our paper. 

\item {\bf New assets}
    \item[] Question: Are new assets introduced in the paper well documented and is the documentation provided alongside the assets?
    \item[] Answer: \answerYes{} % Replace by \answerYes{}, \answerNo{}, or \answerNA{}.
    \item[] Justification:  The code used in the experiments is included in the supplementary material.

\item {\bf Crowdsourcing and research with human subjects}
    \item[] Question: For crowdsourcing experiments and research with human subjects, does the paper include the full text of instructions given to participants and screenshots, if applicable, as well as details about compensation (if any)? 
    \item[] Answer: \answerNA{} % Replace by \answerYes{}, \answerNo{}, or \answerNA{}.
    \item[] Justification: \answerNA{}

\item {\bf Institutional review board (IRB) approvals or equivalent for research with human subjects}
    \item[] Question: Does the paper describe potential risks incurred by study participants, whether such risks were disclosed to the subjects, and whether Institutional Review Board (IRB) approvals (or an equivalent approval/review based on the requirements of your country or institution) were obtained?
    \item[] Answer: \answerNA{} % Replace by \answerYes{}, \answerNo{}, or \answerNA{}.
    \item[] Justification: \answerNA{}

\item {\bf Declaration of LLM usage}
    \item[] Question: Does the paper describe the usage of LLMs if it is an important, original, or non-standard component of the core methods in this research? Note that if the LLM is used only for writing, editing, or formatting purposes and does \emph{not} impact the core methodology, scientific rigor, or originality of the research, declaration is not required.
    %this research? 
    \item[] Answer: \answerYes{} % Replace by \answerYes{}, \answerNo{}, or \answerNA{}.
    \item[] Justification: The LLM usage is described in Section~\ref{sec:llm_usage}.

\end{enumerate}

\end{document}